\pdfoutput=1

\documentclass[11pt]{article}
\usepackage{makecell}
\usepackage{caption}

\usepackage{stackengine}

\usepackage{arydshln}
\usepackage{ACL2023}
\usepackage{times}
\usepackage{latexsym}
\usepackage{enumitem}
\usepackage{array,booktabs}
\usepackage{tabularx}
\usepackage{cuted}
\usepackage{bm}

\newcolumntype{s}{>{\hsize=.35\hsize}X}

\usepackage[T1]{fontenc}

\newcommand{\ci}[1]{\scriptsize $\pm #1$}

\usepackage[utf8]{inputenc}

\usepackage{inconsolata}

\usepackage[ruled]{algorithm2e}
\usepackage{booktabs}
\usepackage{graphicx}
\usepackage{subfigure}
\usepackage{multirow}
\usepackage{multicol}
\usepackage{hyperref}
\usepackage{cite}
\usepackage{makecell}
\usepackage[flushleft]{threeparttable}
\usepackage{amsmath,amssymb,amsfonts}
\usepackage{scalerel,xparse}

\makeatletter
\def\thanks#1{\protected@xdef\@thanks{\@thanks
        \protect\footnotetext{#1}}}
\makeatother

\title{Werewolf Among Us: A Multimodal Dataset for Modeling Persuasion Behaviors in Social Deduction Games}

\author{Bolin Lai* \thanks{* denotes equal contribution.} \\ Georgia Institute of Technology \\  \texttt{bolin.lai@gatech.edu} \And
        Hongxin Zhang* \\ Shanghai Jiao Tong University \\ \texttt{icefox@sjtu.edu.cn} \AND
        Miao Liu* \\ Meta AI \\ \texttt{miaoliu@meta.com} \And
        Aryan Pariani* \\ Georgia Institute of Technology \\ \texttt{apariani3@gatech.edu} \AND
        Fiona Ryan \\ Georgia Institute of Technology \\ \texttt{fkryan@gatech.edu} \And
        Wenqi Jia \\ Georgia Institute of Technology \\ \texttt{wenqi.jia@gatech.edu} \AND
        Shirley Anugrah Hayati \\ University of Minnesota \\ \texttt{sweetpineappleforest@gmail.com} \And
        James M. Rehg \\ Georgia Institute of Technology \\ \texttt{rehg@gatech.edu} \AND
        Diyi Yang \\ Standford University \\ \texttt{diyiy@cs.stanford.edu}
        }

\begin{document}
\maketitle

\begin{abstract}
Persuasion modeling is a key building block for conversational agents. Existing works in this direction are limited to analyzing textual dialogue corpus. We argue that visual signals also play an important role in understanding human persuasive behaviors. In this paper, we introduce the first multimodal dataset for modeling persuasion behaviors. Our dataset includes 199 dialogue transcriptions and videos captured in a multi-player social deduction game setting, $26,647$ utterance level annotations of persuasion strategy, and game level annotations of deduction game outcomes. We provide extensive experiments to show how dialogue context and visual signals benefit persuasion strategy prediction. We also explore the generalization ability of language models for persuasion modeling and the role of persuasion strategies in predicting social deduction game outcomes. Our dataset, code, and models can be found at \url{https://persuasion-deductiongame.socialai-data.org}.
\end{abstract}

\section{Introduction}
As humans, from childhood, we develop the ability to attribute mental belief states to ourselves and others~\citep{premack1978does}. Moreover, we constantly exhibit persuasive behaviors to influence and even reshape the belief states of others during our daily social interactions~\citep{lonigro2017theory}. An automatic system with the ability to understand human persuasion strategies and deduce human belief states may enable more proactive human-computer interaction, and facilitate collaborative decision-making processes.

\begin{figure*}[tbp]
\centering
\includegraphics[width=\linewidth]{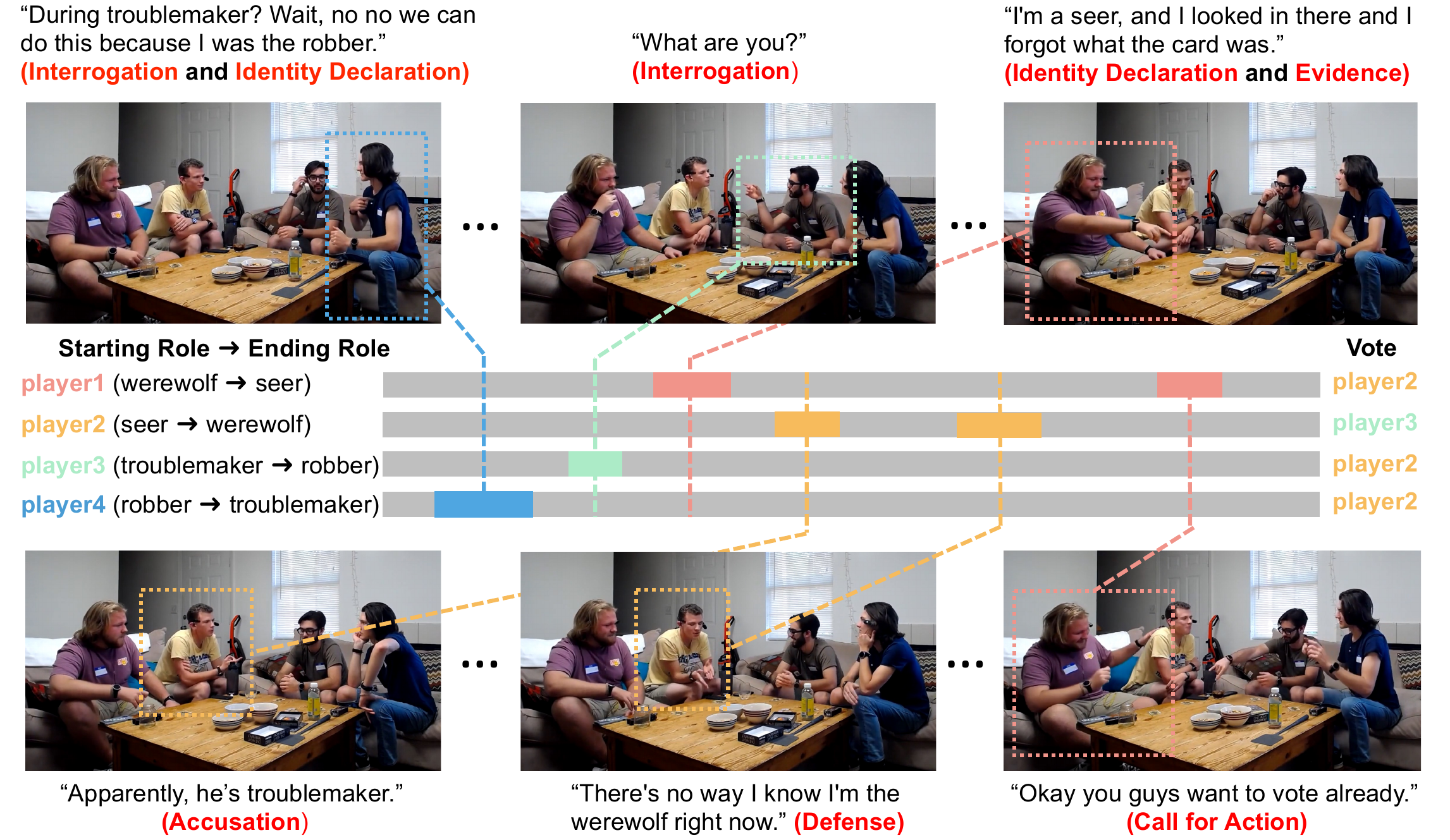}\vspace{-0.5em}
\caption{Demonstration of the six persuasion strategies included in our dataset and the corresponding video. The players are numbered as 1,2,3,4 from left to right. Players' roles might be changed during the game. In this example, player1's and player2's cards were swapped by the troublemaker, and player3's and player4's cards were swapped by the robber. Player 2 voted for player 3 at the end while the others voted for player 2.}
\vspace{-0.5em}
\label{fig:teaser}
\end{figure*}

There have been works on modeling the persuasiveness of arguments and identifying the persuasion strategies utilized on online forums like Reddit, crowd-funding platfroms~\citep{yang-etal-2019-lets,chen2021weakly,atkinson-etal-2019-gets}, and from 1-on-1 dialogues under simulated scenarios through Amazon Mechanical Turk platform ~\citep{wang-etal-2019-persuasion,chawla-etal-2021-casino}.

However, the persuasive behaviors during a naturalistic group discussion with intensive real-time conversation remain unexplored. More importantly, daily human social interaction is multimodal by nature. Both verbal communication (\emph{e.g.} language and audio) and non-verbal communication (\emph{e.g.} gesture and gaze behavior) are essential for analyzing persuasive behavior. 

Moreover, resources for understanding how persuasion strategies affect the deduction outcome during social interaction are missing from the language technologies community.

To bridge these gaps, we introduce \emph{the first multimodal benchmark dataset for modeling persuasive behaviors during multi-players social deduction games}. As shown in Fig.~\ref{fig:teaser}, our dataset is captured in a naturalistic setting where a group of participants plays social deduction games. Our dataset contains both video recordings and the corresponding dialogue transcription. The video data is sourced from both the Ego4D social dataset~\citep{grauman2022ego4d} and YouTube videos. Our dataset also has annotations for persuasion strategy at the utterance level and the voting outcome of each participant during the social deduction game.

We benchmark our dataset by providing comprehensive experimental results and analyzing the role of the video modality and contextual cues in designing the computational models for persuasion behavior prediction. We also provide results to show how different computational models generalize across different data sources and different games. Our contributions are summarized as follows:\vspace{-0.5em}
\begin{itemize}[leftmargin=*]
    \item We present the first multimodal dataset for persuasion modeling. Our dataset is collected in naturalistic social game scenarios with intensive in-person conversations with multiple players.\vspace{-0.5em}
    \item We conduct comprehensive experiments to show the importance of context and visual signals for persuasion strategy prediction.\vspace{-0.5em} 
    \item We provide additional experimental results to investigate model generalization on the persuasion modeling task and discuss how persuasion strategy influences the game voting outcome.
\end{itemize}

\section{Related Work}
\noindent\textbf{Persuasive Behaviors Understanding}. A few previous works introduce datasets for the computational modeling of persuasion~\citep{yang-etal-2019-lets,chen2021weakly,chawla-etal-2021-casino,wang-etal-2019-persuasion,luu-etal-2019-measuring,atkinson-etal-2019-gets}. As summarized in Table~\ref{table:relatedworks}, existing works collect the persuasion language data from the online platform where real-time communication is not available. Moreover, these datasets mainly contain 1-on-1 conversations and lack conversations among multiple speakers. In contrast to these prior efforts, our dataset targets capturing persuasive behaviors in a social group (consisting of 4 to 6 participants). And our work is the first to incorporate visual modality for studying persuasive behaviors.

\begin{table}
\setlength{\tabcolsep}{0.5pt}
\setlength{\extrarowheight}{2pt}
\centering
\footnotesize

\begin{tabular}{c|ccc}
\toprule
Prior Works      &Interaction &Modalities & Setting  \\ \hline 
\citep{yang-etal-2019-lets}   & Online            &Text &Loan \\ 
\citep{chen2021weakly}   & Online            &Text &Request \\ 
\citep{chawla-etal-2021-casino}   & 1 on 1            &Text &Negotiation \\ 
\citep{wang-etal-2019-persuasion}   & 1 on 1            &Text &Charity \\ 
\citep{luu-etal-2019-measuring} & 1 on 1            &Text &Debate \\ 
\citep{atkinson-etal-2019-gets}& Online            &Text &Reddit\\\hline
\multirow{2}{*}{\begin{tabular}[c|]{@{}c@{}}Ours\end{tabular}}    & \multirow{2}{*}{\begin{tabular}[c|]{@{}c@{}}Group \\Discussion\end{tabular}}     
&\multirow{2}{*}{\begin{tabular}[c]{@{}c@{}}Text+Video+\\Audio\end{tabular}} &\multirow{2}{*}{\begin{tabular}[c|]{@{}c@{}}Deduction \\Game\end{tabular}}\\
&   &   &     \\     
\bottomrule
\end{tabular}
\caption{Previous datasets for computational persuasion.}\vspace{-0.5em}
\label{table:relatedworks}
\end{table}

\noindent\textbf{Multimodal Social Interaction}.\ A rich set of literature has addressed the problem of multimodal sentiment analysis. We refer to a recent survey~\citep{kaur2022multimodal} for a more detailed discussion on this topic. The Ego4D social benchmark~\citep{grauman2022ego4d} includes the tasks of identifying who is looking at and talking to the camera wearer using video and audio. Another relevant work~\citep{bara-etal-2021-mindcraft} adopts a multimodal approach to understanding the dialogue behavior in a simulated setting. In contrast, we address the challenging tasks of predicting the persuasion strategy and the deduction outcome from a naturalistic conversation, which requires a richer understanding of high-level social behaviors.

\noindent\textbf{Computational Modeling of Deduction Games}.\ Prior works have investigated computational models for social deduction games. One stream of work seeks to analyze strategies and develop AI agents that play deduction games using a game theory approach ~\citep{nakamura2016constructing, serrino2019finding, chuchro2022training, braverman2008mafia, bi2016human}. These works focus on models of the state of the game alone and do not address understanding the dialogue and persuasive behaviors that often occur while playing. More relevantly, Chittaranjan and Hung \citep{chittaranjan2010you} developed a model for predicting Werewolf game outcomes from player speaking and interrupting behaviors. Recently, Bakhtin et al.~\citep{cicero} introduced a game agent--CICERO, that achieves human-level performance in the Diplomacy game by leveraging a language model with planning and reinforcement learning algorithms. In contrast to these prior works, we present the first work of understanding persuasive behaviors in a group setting from a multimodal perspective.

\section{Dataset}

\begin{table*}[h]
\setlength{\extrarowheight}{2pt}
\centering
\footnotesize
\begin{tabular}{c|c|c|c|c|c|c|c}
\toprule
\multirow{2}{*}{\begin{tabular}[c|]{@{}c@{}}\textbf{Label}\end{tabular}} & \multirow{2}{*}{\begin{tabular}[c|]{@{}c@{}}\textbf{Example}\end{tabular}} & \multicolumn{3}{c|}{\textbf{Ego4D}} & \multicolumn{3}{c}{\textbf{YouTube}} \\ 
\cline{3-8}
{} & {} & \textbf{Count} & \textbf{AUL} & $\boldsymbol{\alpha}$ & \textbf{Count} & \textbf{AUL} & $\boldsymbol{\alpha}$ \\
\hline
\multirow{2}{*}{\begin{tabular}[c|]{@{}c@{}}Identity \\ Declaration\end{tabular}} & \multirow{2}{*}{\begin{tabular}[c|]{@{}c@{}}``I'll just come out and say I was a villager,\\ so I have no idea what's going on.''\end{tabular}} & {\multirow{2}{*}{\begin{tabular}[c|]{@{}c@{}}293\end{tabular}}} & {\multirow{2}{*}{\begin{tabular}[c|]{@{}c@{}}9.87\end{tabular}}} & {\multirow{2}{*}{\begin{tabular}[c|]{@{}c@{}}0.90\end{tabular}}} & {\multirow{2}{*}{\begin{tabular}[c|]{@{}c@{}}1066\end{tabular}}} & {\multirow{2}{*}{\begin{tabular}[c|]{@{}c@{}}10.43\end{tabular}}} & {\multirow{2}{*}{\begin{tabular}[c|]{@{}c@{}}0.87\end{tabular}}} \\ 
& & & & & & &\\
\hline
\multirow{2}{*}{\begin{tabular}[c|]{@{}c@{}}Accusation\end{tabular}} & \multirow{2}{*}{\begin{tabular}[c|]{@{}c@{}}``So James might be the werewolf.''\end{tabular}} & \multirow{2}{*}{\begin{tabular}[c|]{@{}c@{}}669\end{tabular}} & {\multirow{2}{*}{\begin{tabular}[c|]{@{}c@{}}11.28\end{tabular}}} & \multirow{2}{*}{\begin{tabular}[c|]{@{}c@{}}0.74\end{tabular}} & \multirow{2}{*}{\begin{tabular}[c|]{@{}c@{}}2830\end{tabular}} & {\multirow{2}{*}{\begin{tabular}[c|]{@{}c@{}}11.06\end{tabular}}} & \multirow{2}{*}{\begin{tabular}[c|]{@{}c@{}}0.67\end{tabular}} \\
& & & & & & &\\
\hline
\multirow{2}{*}{\begin{tabular}[c|]{@{}c@{}}Interrogation\end{tabular}} & \multirow{2}{*}{\begin{tabular}[c|]{@{}c@{}}``Who did you rob?''\end{tabular}} & \multirow{2}{*}{\begin{tabular}[c|]{@{}c@{}}695\end{tabular}} & {\multirow{2}{*}{\begin{tabular}[c|]{@{}c@{}}7.56\end{tabular}}} & \multirow{2}{*}{\begin{tabular}[c|]{@{}c@{}}0.80\end{tabular}} & \multirow{2}{*}{\begin{tabular}[c|]{@{}c@{}}3407\end{tabular}} & {\multirow{2}{*}{\begin{tabular}[c|]{@{}c@{}}7.66\end{tabular}}} & \multirow{2}{*}{\begin{tabular}[c|]{@{}c@{}}0.90\end{tabular}} \\
& & & & & & &\\
\hline
{\multirow{2}{*}{\begin{tabular}[c|]{@{}c@{}}Call for Action\end{tabular}}}  & {\multirow{2}{*}{\begin{tabular}[c|]{@{}c@{}}``We shouldn't vote to not kill anyone. \\ And then there could also be no werewolf.''\end{tabular}}} & {\multirow{2}{*}{\begin{tabular}[c|]{@{}c@{}}236\end{tabular}}} & {\multirow{2}{*}{\begin{tabular}[c|]{@{}c@{}}9.99\end{tabular}}} & {\multirow{2}{*}{\begin{tabular}[c|]{@{}c@{}}0.78\end{tabular}}} & {\multirow{2}{*}{\begin{tabular}[c|]{@{}c@{}}1163\end{tabular}}} & {\multirow{2}{*}{\begin{tabular}[c|]{@{}c@{}}9.53\end{tabular}}} & {\multirow{2}{*}{\begin{tabular}[c|]{@{}c@{}}0.71\end{tabular}}} \\
& & & & & & &\\
\hline
{\multirow{2}{*}{\begin{tabular}[c|]{@{}c@{}}Defense\end{tabular}}} & {\multirow{2}{*}{\begin{tabular}[c|]{@{}c@{}}``I think that you accused me of \\ being a Werewolf very quickly.''\end{tabular}}} & {\multirow{2}{*}{\begin{tabular}[c|]{@{}c@{}}570\end{tabular}}} & {\multirow{2}{*}{\begin{tabular}[c|]{@{}c@{}}10.04\end{tabular}}} & {\multirow{2}{*}{\begin{tabular}[c|]{@{}c@{}}0.62\end{tabular}}} &  {\multirow{2}{*}{\begin{tabular}[c|]{@{}c@{}}2696\end{tabular}}} & {\multirow{2}{*}{\begin{tabular}[c|]{@{}c@{}}9.75\end{tabular}}} & {\multirow{2}{*}{\begin{tabular}[c|]{@{}c@{}}0.80\end{tabular}}} \\
& & & & & & &\\
\hline
\multirow{2}{*}{\begin{tabular}[c|]{@{}c@{}}Evidence\end{tabular}} & \multirow{2}{*}{\begin{tabular}[c|]{@{}c@{}}``If you swapped these two, \\ he is not the werewolf.''\end{tabular}} & \multirow{2}{*}{\begin{tabular}[c|]{@{}c@{}}489\end{tabular}} & {\multirow{2}{*}{\begin{tabular}[c|]{@{}c@{}}11.45\end{tabular}}} & \multirow{2}{*}{\begin{tabular}[c|]{@{}c@{}}0.75\end{tabular}} & \multirow{2}{*}{\begin{tabular}[c|]{@{}c@{}}1740\end{tabular}} & {\multirow{2}{*}{\begin{tabular}[c|]{@{}c@{}}9.80\end{tabular}}} & \multirow{2}{*}{\begin{tabular}[c|]{@{}c@{}}0.60\end{tabular}} \\
& & & & & & &\\
\bottomrule
\end{tabular}
\caption{Utterance-level persuasion strategy annotations. AUL refers to the average utterance length in terms of the number of words in an utterance and $\alpha$ refers
to Krippendorff’s alpha.}\vspace{-0.5em}
\label{table:persuasion_strategy_annotation}
\end{table*} 

\subsection{Data collection}
To benchmark the generalization ability of the computational models, we collect our data from different sources, as detailed next. This work has been approved by the Institutional Review Board.

\paragraph{Ego4D Dataset} We first leverage a subset of Ego4D Social dataset~\citep{grauman2022ego4d} for our study. This subset captures videos of groups of participants playing social deduction games. This subset contains $7.3$ hours of videos with 40 games of One Night Ultimate Werewolf and $8$ games of The Resistance: Avalon. We refer to Appendix~\ref{app:rule} for rules of these two games. Note that the Avalon data has a relatively small scale, and therefore is only used to evaluate the cross-domain generalization ability of our models.  To ensure all participants are visible in the frame, we use third-person videos instead of the first-person videos from Ego4D for visual representation learning and transcription.

\paragraph{YouTube Video} We retrieve the top search results YouTube videos under the keywords of “one night ultimate werewolf” and “ultimate werewolf”. We further select a final set of $14.8$ hours of videos with 151 clips of completed games that adopt the same game setup as the Ego4D dataset and have fully visible game outcomes. We will release the YouTube URLs for the selected videos.

\subsection{Data Annotation}

\paragraph{Video Annotation} Most Ego4D and YouTube videos contain multiple games. Therefore, we first annotate the starting time (when the game narration voice starts) and the ending time (right before the voting stage) of each game. We then ask the annotators to look through each game clip and annotate the starting role, ending role, and the voting outcome of each player.

\paragraph{Transcription} We use an automatic transcription service \emph{rev.com} to generate the transcript of each game clip. We ask  annotators to carefully examine the alignment of videos and transcripts, and manually correct any errors in the transcripts. Please refer to Appendix~\ref{app:transcription_interface} for more details.

\paragraph{Persuasion Strategy Annotation}
Inspired by prior psychology studies and other works on predicting persuasion strategies~\citep{chawla-etal-2021-casino,carlile-etal-2018-give,yang-etal-2019-lets,chen2021weakly}, we propose six persuasion tactics that are frequently adopted in social deduction games. \vspace{-0.5em}

\begin{itemize}[leftmargin=*]
    \item \textbf{Identity Declaration}: \emph{State one's own role or identity in the game}. This is a game-specific persuasion tactic.\vspace{-0.5em}
    \item \textbf{Accusation}: \emph{Claim someone has a specific identity or strategic behavior}. Accusation, similar to Undervalue-Partner~\citep{chawla-etal-2021-casino}, is a generic proself behavior. \vspace{-0.5em}
    \item \textbf{Interrogation}: \emph{Questions about someone's identity or behavior}.  Interrogation, is a proself strategy related to individual preferences. \vspace{-0.5em}
    \item \textbf{Call for Action}: \emph{Encourage people to take an action during the game}. Call for Action relates to the coordination for persuasion~\citep{chawla-etal-2021-casino}, which is a generic prosocial behavior. \vspace{-0.5em}
    \item \textbf{Defense}: \emph{Defend oneself or someone else against an accusation or defend a game-related argument}. An utterance depicts Defense when the persuader tries to use credentials impacts  earn others' trust or justify their earlier decisions.
    \vspace{-0.5em}
    \item \textbf{Evidence}: \emph{Provide a body of game-related fact or information}. Evidentiality is a general persuasion tactic that has been widely studied in previous works~\citep{carlile-etal-2018-give}.\vspace{-0.5em}
\end{itemize}

Following previous work~\citep{chawla-etal-2021-casino}, we annotate the persuasion strategy at the utterance level. We provide a website annotation tool adopted from \citep{hayati-etal-2020-inspired} for our annotation task (see Appendix~\ref{app:interface} for details) to the annotators. To properly train the annotators, we first ask all three annotators to annotate the same subset of dialogues and compute inter-annotator agreement using the nominal form of krippendorff's alpha~\citep{krippendorff2018content}. We then discuss with the annotators on their disagreements and come up with a general rule to address the disagreements during the annotation process. We repeat the above process until the annotators reached a krippendorff's alpha greater than $0.6$ for each category. In Table~\ref{table:persuasion_strategy_annotation}, we report the per-class Krippendorff's alpha value for the final round of inter-annotator agreement calculation. After the annotator training phase is completed, we ask the three annotators to independently annotate the rest of the Ego4D and YouTube data.

\paragraph{Annotation Statistics}
Our dataset has $5,815$ utterances from the Ego4D data, and $20,832$ utterances from the YouTube data. More than $49.2\%$ of Ego4D utterances are labeled as no strategy because of the naturalistic social setting, while only $37.9\%$ of YouTube utterances are labeled as no strategy since players from the YouTube videos are more proficient at the game and focused more on gameplay. Furthermore, as shown in Table~\ref{table:persuasion_strategy_annotation}, the adopted persuasion strategies have an imbalanced distribution, where ``Accusation'', ``Interrogation'', and ``Defense'' are the most frequent strategies for both the Ego4D and YouTube videos. The annotators are recruited from a Startup Data Platform dedicated to research projects. All annotators are paid hourly at a rate above the federal minimum.

\section{Strategy Prediction}
Given an utterance and its corresponding video segment, we seek to predict the persuasion strategies adopted in the utterance. We first leverage a pre-trained language model~\citep{devlin-etal-2019-bert,roberta} as the text encoder to obtain the utterance embedding, and a vision transformer~\citep{fan2021multiscale} to obtain the visual embedding. We then concatenate the textual and visual features to predict the persuasion strategy. Additionally, we study the impact of textual context by including prior utterances as input. 

\subsection{Methodology}

\begin{figure}[tbp]
\centering
\includegraphics[width=1.0\linewidth]{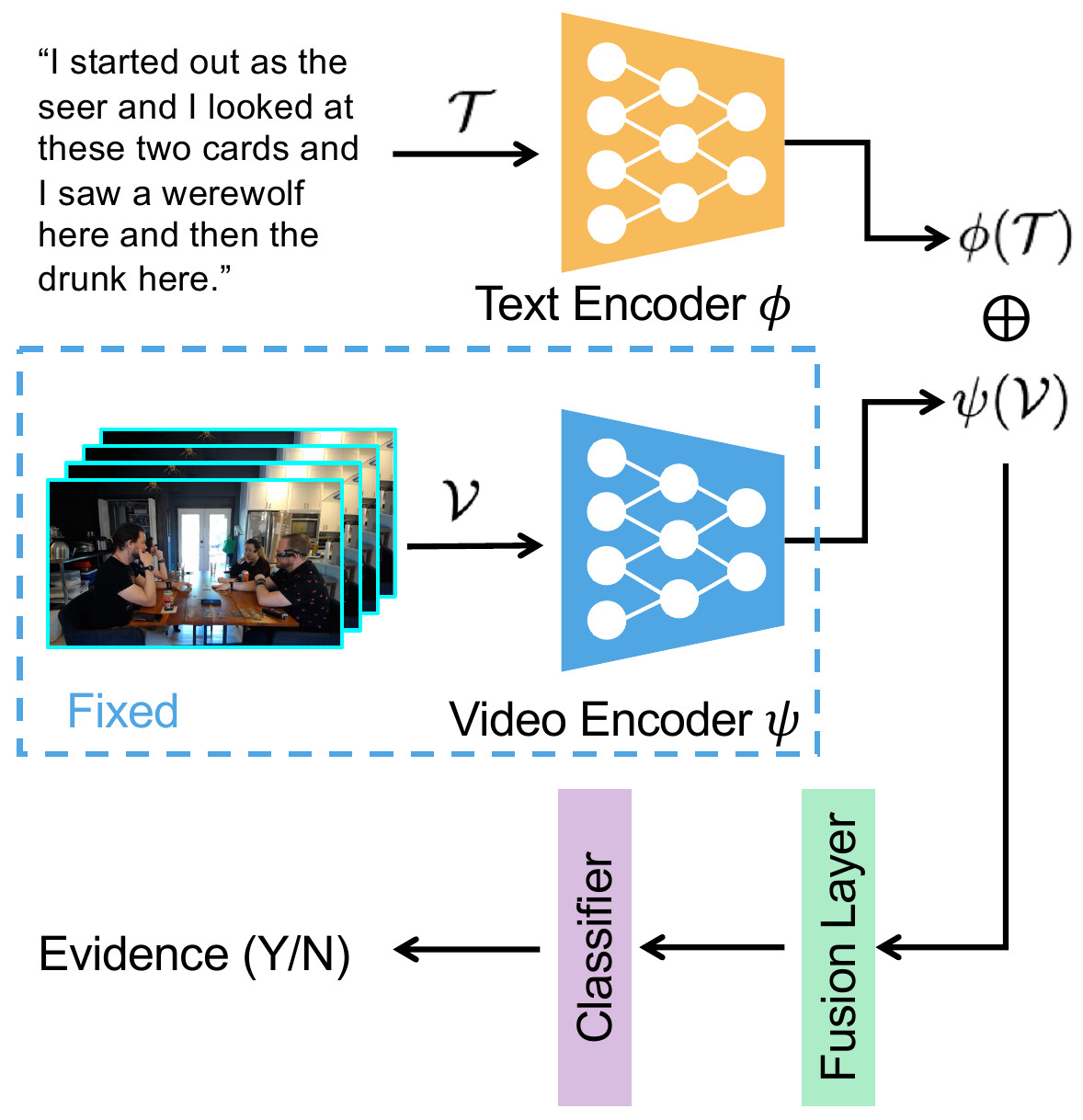}\vspace{-0.5em}
\caption{Architecture of the independent model for each strategy. We fix the parameters in the video encoder and train the other modules end-to-end.  $\oplus$ denotes the concatenation of two features.}\vspace{-0.5em}
\label{fig:indp_model}
\end{figure}

In our dataset, an utterance may be labeled with multiple persuasion strategies. For instance, \emph{``I'm a villager and she is the werewolf.''} is labeled as both \emph{identity declaration} and \emph{accusation}. Therefore, we formulate this task as a binary classification problem for each strategy and consider an utterance as non-strategic if it gets negative labels in all strategies. The most straightforward approach to solve this task would be fine-tuning a pre-trained language model, which is referred to as \emph{Base} model. In addition, we consider the following approaches:\smallskip

\noindent\textbf{Modeling with Context Embedding}. Since some persuasion strategies cannot be easily recognized from one single utterance, we further consider a model with additional context for each utterance. This is denoted as \emph{Base + C}.

\noindent\textbf{Modeling with Video Representation}. We further leverage the non-verbal signals by combining video features with the text representation for persuasion modeling. We directly use a pre-trained Vision Transformer to extract video representations, and fuse video and text representations before feeding them into the classification layer as shown in Fig.~\ref{fig:indp_model}. We refer to this model as \emph{Base + V}.\smallskip

\noindent\textbf{Late fusion of Video and Context}. Finally, we adopt a late fusion model (\emph{Base + C + V})  that incorporates both video features and context cues for persuasion strategy prediction.

\subsection{Model Details}
We perform our experiments using both \textbf{BERT}~\citep{devlin2018bert} and \textbf{RoBERTa}~\citep{roberta} as backbones for the text encoder. We use \texttt{bert-base-uncased} and \texttt{roberta-base} models from Huggingface \citep{wolf-etal-2020-transformers} in our implementation. We adopt MViT-B-24~\citep{fan2021multiscale} pretrained on Kinetics-400~\citep{kay2017kinetics} as the video encoder. Moreover, we also implement a multi-task model~\citep{chawla-etal-2021-casino} as an additional baseline, referred to as \textbf{MT-BERT}. Context and video features are incorporated into MT-BERT in the same way as BERT and RoBERTa.\smallskip

\noindent \textbf{Base Model}. For base models, we obtain the textual input $\mathcal{T}$ from the current utterance only. Then we input $\mathcal{T}$ into a text encoder $\phi$ followed by a classifier to get the strategy prediction.\smallskip

\noindent\textbf{Base + C}.\ We first concatenate the $k$ previous utterances $C_1, C_2, \cdots, C_k$ with an \emph{[EOS]} token to get context $C$, and then concatenate this with the current utterance $U$ using an \emph{[SEP]} token to get the final input $\mathcal{T}$. Formally, we have
\begin{align}
    & C = C_1\ \text{[EOS]}\ C_2\ \text{[EOS]} \cdots C_k, \\
    & \mathcal{T} = C\ \text{[SEP]}\ U.
\end{align}

\begin{figure*}[tbp]
\centering
\includegraphics[width=0.9\linewidth]{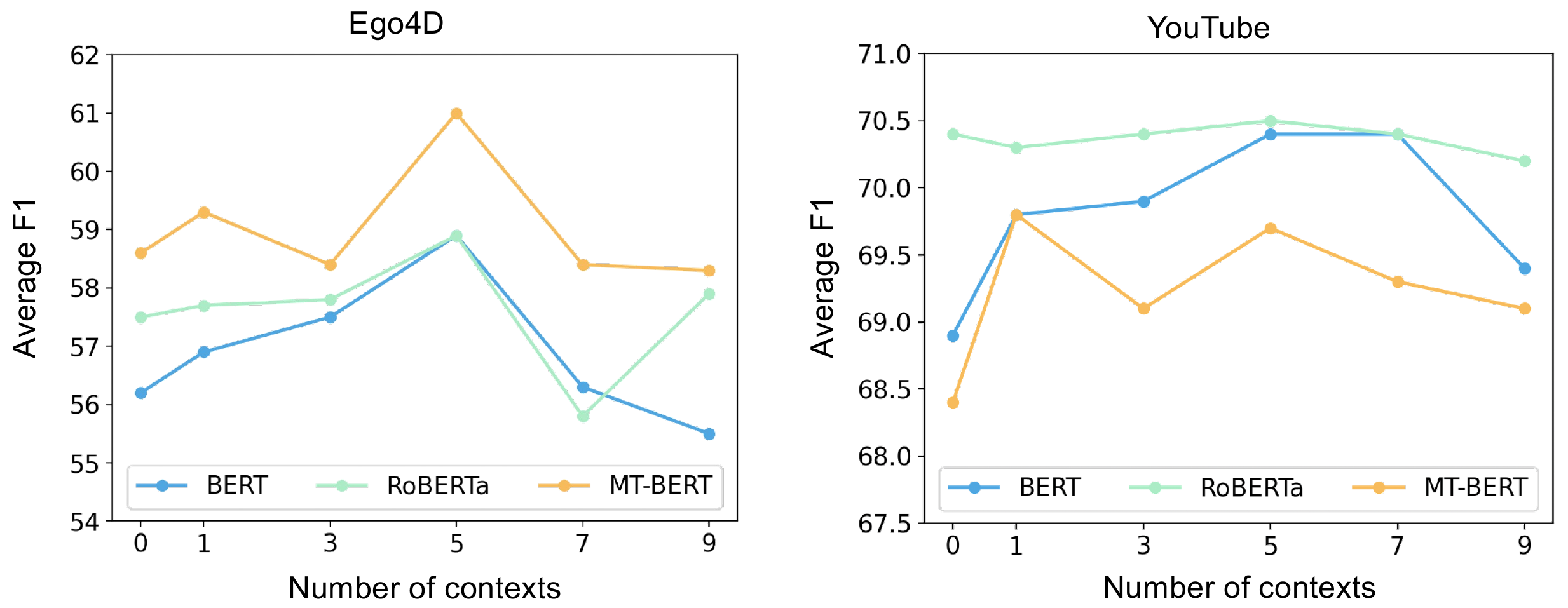}\vspace{-0.5em}
\caption{Ablation study of adopting different context lengths for persuasion strategy prediction.}\vspace{-0.5em}
\label{fig:context_ablation}
\end{figure*}

\noindent\textbf{Base + V}. 
We use video encoder $\psi$ to extract visual representation $\psi(\mathcal{V})$ of the corresponding video clip $\mathcal{V}$. During training, video features are concatenated with the text representation $\phi(\mathcal{T})$ and fed into a fusion layer, which uses a linear mapping function $W_F^T$ and an activation function $Tanh(\cdot)$. Finally, we apply a linear classifier $W_P^T$ to obtain the prediction logits, which can be formulated as
\begin{equation}
    logits = W_P^T\cdot Tanh\left(W_F^T\cdot(\phi(\mathcal{T}) \oplus\psi(\mathcal{V}))\right),
\end{equation}

\noindent where $\oplus$ denotes the concatenation of two vectors. Note that we fix the parameters of the video encoder during training. Please refer to Appendix~\ref{app:video} for more details on visual representation extraction.\smallskip
    
\noindent\textbf{Base + C + V}. We further late fuse \textbf{+ C} with \textbf{+ V}. Formally, we denote the probability predictions of the two models after softmax as $P_{C}$ and $P_{V}$. Then the output after linear combination is formulated as
\begin{equation}
    P_{C,V} = (1-\lambda) P_C + \lambda P_V,
\end{equation}
where $\lambda$ is a scalar that balances $P_{C}$ and $P_{V}$.\vspace{-0.3em}

\subsection{Training Details}
All models are trained using cross-entropy loss. For training hyper-parameters, we do a grid search of learning rates in $\{1e-5, 3e-5, 5e-5\}$ and batch sizes in $\{16, 8\}$ for \emph{Base} models. We then fix the optimal hyper-parameters for subsequent models incorporating context or videos. We train all models with the optimal learning rates and batch sizes for 10 epochs using AdamW~\citep{loshchilov2017decoupled} as the optimizer. We run all the experiments with three random seeds and report the average score and the standard deviation.

\subsection{Experiment Results}

\noindent\textbf{Evaluation Metrics}.\ Following~\citep{chawla-etal-2021-casino}, we report the F1 score for each category, the average F1 score of all categories, and Joint Accuracy. Note that the prediction is considered as correct when all the categories are predicted correctly in Joint Accuracy.\smallskip

\noindent\textbf{Ablations on Additional Context}.\ We first present a systematic ablation study of how incorporating textual context may improve the performance of persuasion strategy prediction. Specifically, we feed a fixed length of previous utterances together with the current utterance into the backbone language encoders for classification. As shown in Fig.~\ref{fig:context_ablation}, the additional context can boost the performance of all baseline models. However, setting the context length for too long may confuse the model, especially for categories that can be reliably predicted with the current utterance (\emph{e.g.} Identity Declaration, and Interrogation). We present the per-class performance in Appendix~\ref{app:per-class}. Our empirical finding is that a context length of $5$ can consistently improve the performance of all three baseline models. Therefore, we adopt a context length of $5$ as a default setting for the rest of our experiments. \smallskip

\begin{table*}[t]
\footnotesize 
\centering
{
\setlength{\tabcolsep}{3.2pt}
\setlength{\extrarowheight}{2pt}

\begin{tabular}{c|c|cccccc|cc}
\toprule
&Method &Identity	&Accusation	&Interrogation	&Call for Action	&Defense	&Evidence	&Avg F1	&Joint-A       \\ 
\hline 
\multirow{12}{*}{\rotatebox[origin=c]{90}{Ego4D}}  
&\makecell{BERT}  & $82.6$\ci{1.1}  & $48.8$\ci{4.8}  & $82.8$\ci{0.2}  & $39.4$\ci{9.6}  & $29.3$\ci{5.5}  & $54.2$\ci{2.5}  & $56.2$\ci{2.5}  & $65.1$\ci{1.6}  \\
&\makecell{BERT + C} & $79.9$\ci{1.6}  & $52.0$\ci{3.3}	 & $81.0$\ci{1.1}  & $49.5$\ci{3.2}	 & $33.8$\ci{0.5}  & $57.1$\ci{1.6}	 & $58.9$\ci{0.6}  & $65.0$\ci{0.2}  \\
&\makecell{BERT + V} & $81.5$\ci{3.5}  & $52.1$\ci{1.9}  & $83.3$\ci{1.6}  & $42.4$\ci{3.8}  & $28.4$\ci{5.1}  & $52.8$\ci{1.0}  & $56.7$\ci{1.2}  & $64.5$\ci{1.2}  \\ 
&\makecell{BERT + C + V}   & {\fontseries{b}\selectfont 84.5}\ci{4.6}  & $52.8$\ci{2.0}  & $82.7$\ci{0.4}  & $47.3$\ci{3.4}  & $34.5$\ci{1.7}  & $54.9$\ci{1.1}  & $59.4$\ci{1.6}  & {\fontseries{b}\selectfont 66.5}\ci{0.3}  \\ 
\cdashline{2-10}
&\makecell{RoBERTa}   & $81.7$\ci{2.6}  & $51.7$\ci{0.9}  & $83.4$\ci{0.9}  & $43.3$\ci{8.7}  & $33.1$\ci{2.2}  & $51.7$\ci{2.1}  & $57.5$\ci{1.4}  & $63.4$\ci{0.5} \\
&\makecell{RoBERTa + C}  & $81.5$\ci{0.7}  & {\fontseries{b}\selectfont 59.4}\ci{2.4}  & $83.5$\ci{1.1}  & $43.7$\ci{3.7}  & $33.0$\ci{3.1}  & $52.4$\ci{2.9}  & $58.9$\ci{1.2}  & $64.6$\ci{0.7} \\
&\makecell{RoBERTa + V}  & $79.8$\ci{0.6}  & $51.4$\ci{1.0}  & $82.8$\ci{2.1}  & $50.1$\ci{5.3}  & $31.3$\ci{3.1}  & $54.6$\ci{3.2}  & $58.3$\ci{0.7}  & $64.0$\ci{0.9} \\
&\makecell{RoBERTa + C + V}   & $82.7$\ci{0.2}  & $58.5$\ci{2.3}  & $83.8$\ci{1.2}  & $46.1$\ci{4.5}  & $35.4$\ci{3.4}  & $53.4$\ci{3.3}  & $60.0$\ci{0.8}  & $66.1$\ci{0.9} \\ 
\cdashline{2-10}
&\makecell{MT-BERT}   & $80.9$\ci{1.3}  & $51.5$\ci{3.3}  & $83.0$\ci{1.3}  & {\fontseries{b}\selectfont 56.6}\ci{2.3}  & $25.9$\ci{2.0}  & $53.6$\ci{1.3}  & $58.6$\ci{0.3}  & $65.5$\ci{0.8}  \\
&\makecell{MT-BERT + C}     & $79.8$\ci{2.2}  & $54.4$\ci{0.8}  & $83.2$\ci{0.7}  & $50.8$\ci{7.2}	 &{\fontseries{b}\selectfont 36.5}\ci{2.8}  & {\fontseries{b}\selectfont 61.5}\ci{2.2}	 & {\fontseries{b}\selectfont 61.0}\ci{1.1}  & $66.3$\ci{1.4}  \\
&\makecell{MT-BERT + V} & $79.9$\ci{1.6}  & $51.9$\ci{0.8}  &{\fontseries{b}\selectfont 84.8}\ci{2.4}  & $53.9$\ci{4.5}  & $35.4$\ci{2.2}  & $53.3$\ci{1.0}  & $59.8$\ci{0.7}  & $62.1$\ci{3.4}  \\ 
&\makecell{MT-BERT + C + V}   & $80.7$\ci{1.9}  & $55.2$\ci{0.9}  & $83.6$\ci{0.6}  & $50.0$\ci{0.8} & $36.1$\ci{2.7}  & $60.5$\ci{1.0}  & {\fontseries{b}\selectfont 61.0}\ci{0.3}  & $66.3$\ci{1.0}   \\ 
\hline
\multirow{12}{*}{\rotatebox[origin=c]{90}{YouTube}}  
&\makecell{BERT}   & $80.2$\ci{1.6}  & $64.7$\ci{1.1}  & $89.6$\ci{0.4}  & $77.2$\ci{2.5}  & $43.5$\ci{1.0}  & $58.3$\ci{0.7}  & $68.9$\ci{0.0}  & $64.6$\ci{0.8}  \\
&\makecell{BERT + C}   & $82.6$\ci{0.7}  & $66.7$\ci{1.0}  & $89.6$\ci{1.5}  & $78.1$\ci{2.4}  & $45.7$\ci{1.1}  & $59.7$\ci{1.1}  & $70.4$\ci{0.3}  & $64.4$\ci{1.0}  \\
&\makecell{BERT + V}   & $82.4$\ci{0.5}  & $65.4$\ci{1.4}  & $89.7$\ci{0.1}  & $78.0$\ci{0.8}  & $45.3$\ci{2.8}  & $58.4$\ci{1.3}  & $69.9$\ci{0.4}  & $66.2$\ci{0.5}  \\ 
&\makecell{BERT + C + V}   & $83.6$\ci{0.1}  & $67.2$\ci{1.2}  & {\fontseries{b}\selectfont 90.2}\ci{1.0}  & $78.5$\ci{1.6}  & $46.6$\ci{1.1}  & $59.9$\ci{1.0}  & $71.0$\ci{0.2}  & $66.7$\ci{0.5}  \\ 
\cdashline{2-10}
&\makecell{RoBERTa}   & $84.3$\ci{0.1}  & $67.2$\ci{0.6}  & $89.4$\ci{0.1}  & $78.2$\ci{0.8}  & $44.3$\ci{0.4}  & $59.0$\ci{1.7}  & $70.4$\ci{0.2}  & $64.8$\ci{0.7} \\
&\makecell{RoBERTa + C}   & $82.4$\ci{0.3}  & $67.0$\ci{1.1}  & {\fontseries{b}\selectfont 90.2}\ci{0.0}  & $77.1$\ci{1.0}  & $46.1$\ci{0.7}  & $59.9$\ci{0.7}  & $70.5$\ci{0.3}  & $64.7$\ci{0.6}  \\
&\makecell{RoBERTa + V}  & $83.4$\ci{0.4}  & $66.4$\ci{0.3}  & $89.5$\ci{0.1}  & {\fontseries{b}\selectfont 78.7}\ci{2.0}  & $46.6$\ci{0.6}  & $59.0$\ci{1.0}  & $70.6$\ci{0.1}  & $65.3$\ci{1.2}  \\ 
&\makecell{RoBERTa + C + V}  & $83.7$\ci{0.6}  & $67.4$\ci{0.4}  & $89.8$\ci{0.3}  & $78.5$\ci{1.2}  & {\fontseries{b}\selectfont 48.2}\ci{0.7}  & $60.4$\ci{0.8}  & {\fontseries{b}\selectfont 71.3}\ci{0.2}  & $66.4$\ci{0.7}  \\ 
\cdashline{2-10}
&\makecell{MT-BERT} & $80.7$\ci{0.4}  & $65.1$\ci{1.5}  & $88.5$\ci{0.8}  & $76.2$\ci{2.2}  & $42.3$\ci{1.5}  & $57.4$\ci{1.3}  & $68.4$\ci{0.3}  & $65.6$\ci{1.1} \\
&\makecell{MT-BERT + C} & $83.1$\ci{1.1}  & $65.0$\ci{1.5}  & $90.1$\ci{0.3}  & $74.6$\ci{2.4} & $46.5$\ci{0.8}  & $59.2$\ci{0.3}  & $69.7$\ci{0.6}  & $66.7$\ci{0.5} \\
&\makecell{MT-BERT + V} & $82.8$\ci{0.6}  & {\fontseries{b}\selectfont 68.5}\ci{1.0}  & $89.3$\ci{0.7}  & $75.6$\ci{2.8} & $47.8$\ci{0.3}  & $59.6$\ci{0.8}  & $70.6$\ci{0.8}  & $66.9$\ci{0.4}  \\ 
&\makecell{MT-BERT + C + V} & {\fontseries{b}\selectfont 84.4}\ci{0.6}  & $68.4$\ci{1.0}  & $89.5$\ci{0.6}  & $76.5$\ci{2.1} & $47.3$\ci{0.5}  & {\fontseries{b}\selectfont 60.6}\ci{0.2}  & $71.1$\ci{0.5}  & {\fontseries{b}\selectfont 68.1}\ci{0.2}  \\
\bottomrule
\end{tabular}}
\caption{Experimental Results on incorporating visual features for persuasion strategy prediction. We train an independent model for each category using BERT and RoBERTa backbones. Additionally, we also use the off-the-shelf Multi-Task BERT model (MT-BERT)~\citep{chawla-etal-2021-casino} to jointly predict all categories. }\vspace{-0.5em}
\label{table:videofeature}
\end{table*}

\begin{table}[t]
\centering
{
\footnotesize 
\begin{tabular}{ccccc}
\toprule
\multirow{2}{*}{Setting}          
&\multicolumn{2}{c}{Ego4D} &\multicolumn{2}{c}{YouTube} \\
\cmidrule(lr){2-3} \cmidrule(lr){4-5} 
 & Avg F1  & Joint-A      & Avg F1 & Joint-A  \\
\midrule
\makecell{Majority}   & 0 & 52.5 & 0 & 38.8 \\
\midrule
\makecell{Zero-Shot}   & 35.4 & 58.5 & 40.3 & 52.0 \\
\makecell{One-Shot}    & 40.7 & 56.3 & 47.2 & 53.2 \\						     
\makecell{Five-Shot}   & 47.0 & 59.7 & 49.6 & 53.7 \\ 
\bottomrule
\end{tabular}}
\caption{GPT-3 results on Ego4D and YouTube data.}\vspace{-0.5em}
\label{table:gpt}
\end{table}

\noindent\textbf{Modeling with Video Representation}.\ We further study how incorporating video representation improves the performance of persuasion modeling. The results are summarized in Table~\ref{table:videofeature}. Importantly, video features can improve the BERT model by ~$0.8\%$ on both the Ego4D dataset and YouTube dataset. However, RoBERTa+V only beats the RoBERTa model by $0.2\%$ on the YouTube dataset. This might be because the YouTube dataset has more training data for the RoBERTa model to learn a good representation without video feature embedding. Interestingly, including video features has a larger performance boost on predicting ``Accusation'',  ``Interrogation'',  and ``Call for Action'', which is probably due to the more frequent non-verbal communication (\emph{e.g.} pointing to someone, raising hands, turning the head) during these persuasive behaviors.\smallskip

\begin{figure*}[tbp]
\centering
\includegraphics[width=0.95\linewidth]{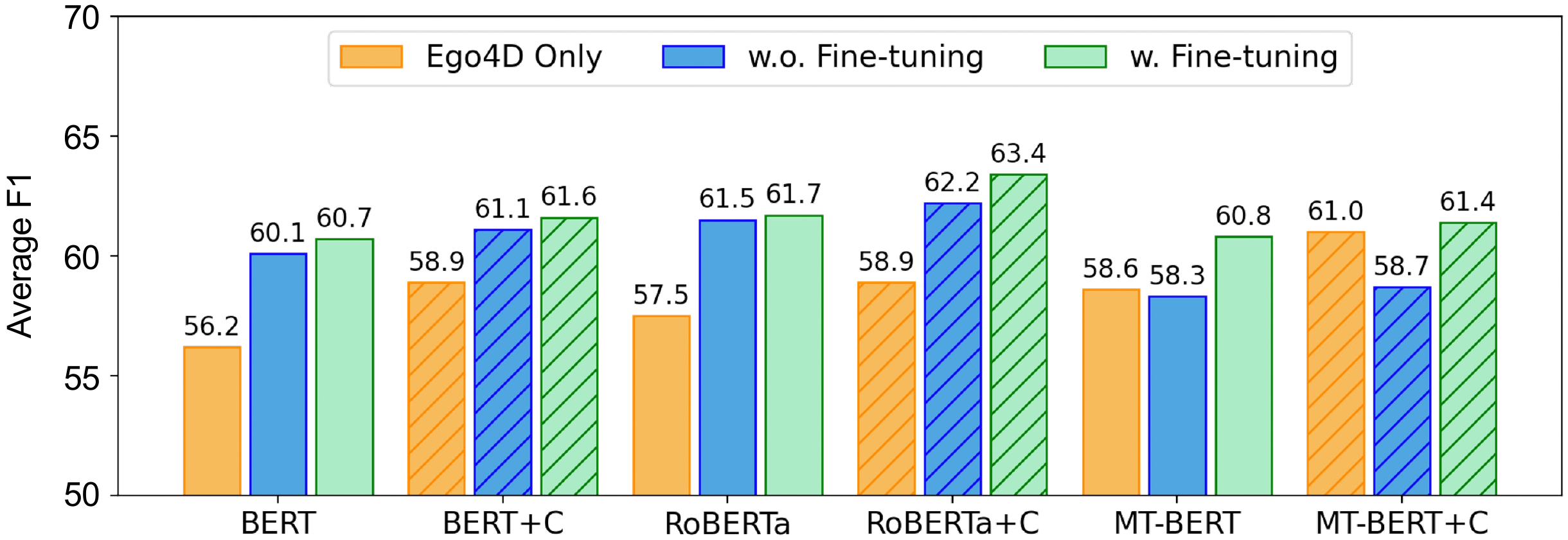}\vspace{-0.5em}
\caption{Data domain generalization experiments. We report the testing performance on the Ego4D dataset using models trained only on YouTube data  (w.o. Fine-tuning), and trained on YouTube data and further fine-tuned with Ego4D data (w. Fine-tuning). We also report the performance (refer to Table~\ref{table:videofeature}) of the models trained only on Ego4D dataset (Ego4D Only) as comparison.}\vspace{-1.0em}
\label{fig:data_generalization}
\end{figure*}

\begin{figure}[tbp]
\centering
\includegraphics[width=1.0\linewidth]{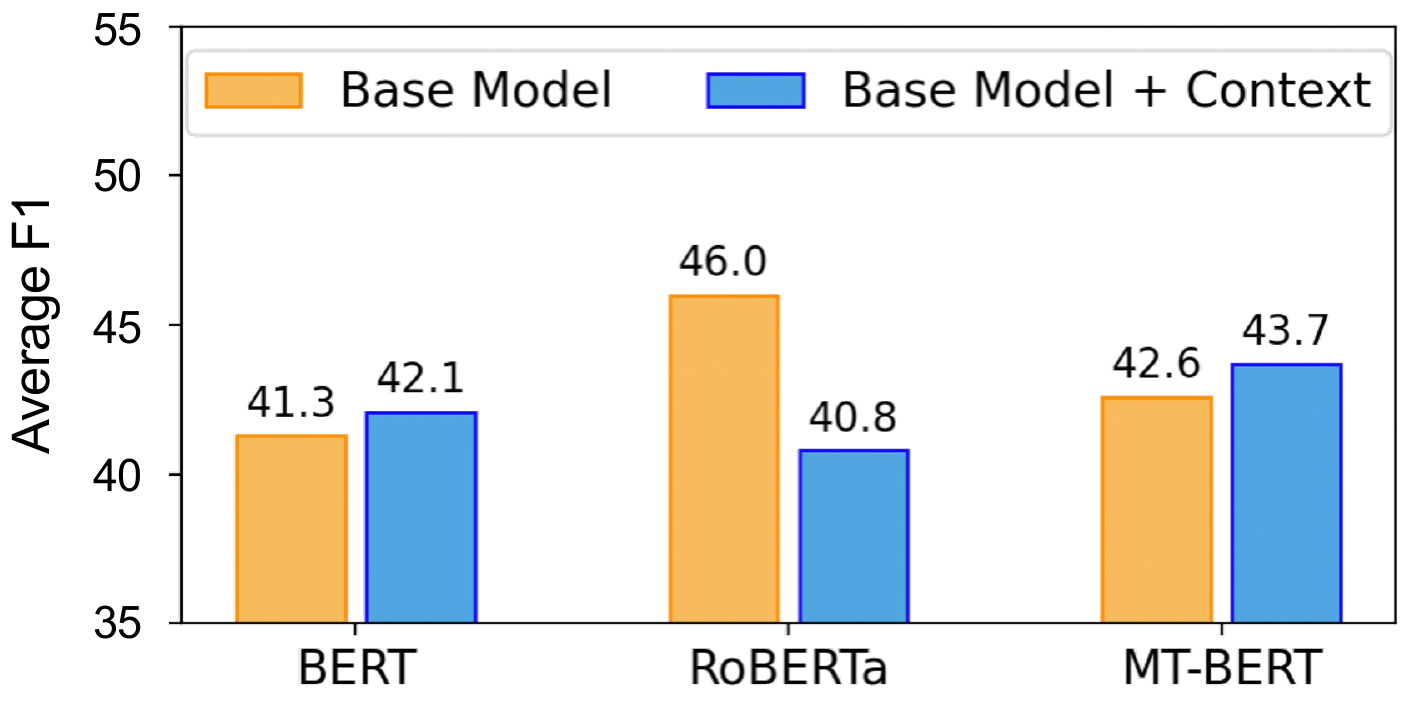}\vspace{-0.5em}
\caption{Game domain generalization experiments. We report the testing performance on Ego4D Avalon data using models trained only on Ego4D Werewolf data.}\vspace{-1.0em}
\label{fig:game_generalization}
\end{figure}

\noindent\textbf{Off-the-shelf GPT-3 Inference}.\
Prompting Large Language Models off-the-shelf to solve NLP tasks has received increasing attention \citep{GPT-3}. Here, we experiment with GPT-3-175B on our benchmark under three settings: zero-shot, one-shot and five-shot. Specifically, we use \texttt{text-davinci-002} engine from OpenAI's API\footnote{https://beta.openai.com/} with temperature 0 to produce a deterministic answer. The detailed templates for different settings are shown in Appendix~\ref{app:prompts}. The result is shown in Table~\ref{table:gpt}. Using GPT-3 off-the-shelf achieves a non-trivial performance (Joint-A of 52.0 v.s. 38.8 for majority on YouTube data). Adding more examples further boosts performance, though it is still inferior to the fine-tuned models.\smallskip

\noindent\textbf{Data Domain Generalization}.\ We conduct additional experiments to show the generalization ability of language models on persuasion prediction. To begin with, we use the model trained on the YouTube data to make predictions on the Ego4D Werewolf testing data without any fine-tuning. As shown in Fig.~\ref{fig:data_generalization}, the resulting model achieves better performance than models trained only on Ego4D in most cases, due to the larger amount of available training data from the YouTube dataset. This also suggests that, for the text modality, the domain gap between the Ego4D and the YouTube data is small. We further fine-tune the model trained on the YouTube data with the Ego4D training data, and the resulting model performs even better. These results suggest promise in leveraging the large body of videos available online as a pre-training source for persuasion modeling in naturalistic social interactions. Another take-home message from our experiments is that the multi-task setting (MT-BERT) may compromise the model generalization ability. We also find that including video representation cannot improve the model generalization ability (see more details in Appendix~\ref{app:video}), suggesting that the video modality domain gap between the two data sources is much larger than the text modality.\smallskip

\noindent\textbf{Game Domain Generalization}.\
We also study the model generalization ability on another social game -- Avalon. Werewolf and Avalon are vastly different in the game rules and winning conditions, especially because Werewolf has only one voting round per game, while Avalon has multiple rounds per game. Therefore, the persuasion strategies adopted in Avalon have a different distribution from Werewolf (see Appendix~\ref{app:avalon}). We run inference on the Avalon data using models trained only on the Ego4D Werewolf data without fine-tuning. Results are shown in Figure~\ref{fig:game_generalization}. Despite the large domain gap between the two games, our models achieve decent performance on the Avalon data. However, we find incorporating additional context has marginal performance improvements, and may even compromise the performance of the RoBERTa model. The detailed results of data and game domain generalization are shown in Appendix~\ref{app:domain_generalization}.\vspace{-0.3em}

\section{Game Outcome Deduction} 

\vspace{-0.2em}
In addition to predicting persuasion strategies, we further model the human deduction process by predicting the voting outcomes of each pair of players, \emph{i.e.}, whether player A (\emph{voter}) votes for player B (\emph{candidate}). Therefore, in a game of $n$ players, there are $C_n^2$ (Combinations) of player pairs, corresponding to $P_n^2$ (Permutations) data points. We merge all data points from the Ego4D Werewolf data and YouTube data to enlarge the dataset size, and split the resulting data into 2741/427/827 samples for train/val/test sets. Since each player is only allowed to vote for one player, the resulting data has an imbalanced distribution, with $20.4\%$ of the samples being positive (positive indicates the voter votes for the candidate).\vspace{-0.5em}

\subsection{Method}
For deduction modeling, we encode the input feature with three embeddings: a $7\times1$ vector representing the persuasion strategy distribution (including non-strategy) adopted by the voter; a $7\times1$ vector representing the persuasion strategy distribution adopted by the candidate; and a $12\times1$ one-hot vector representing starting role of the voter (One Night Werewolf has 12 roles in total). Therefore, the input for deduction is a $26\times1$ vector. We use a simple logistic regression model for deduction modeling. To address the class imbalance of positive and negative samples, we train the model with weighted binary classification loss. \vspace{-0.3em}

\subsection{Experiment Results}

Our model achieves an F1 of $32.7\%$ and an AUC of $54.7\%$, outperforming the random prediction, which obtains F1 and AUC of $28.6\%$ and $50.0\%$, respectively. These results show the effectiveness of our proposed method. To analyze the effect of persuasion strategy embedding and role embedding, we consider another baseline model that only takes the persuasion strategy embedding of the paired players as the inputs. The resulting model achieves an F1 of $32.2\%$ and an AUC of $54.6\%$. 
Overall we found that the persuasion strategy embedding is more informative for the deduction of human intention than the role embedding.

\begin{figure}[t]
\centering
\includegraphics[width=1.0\linewidth]{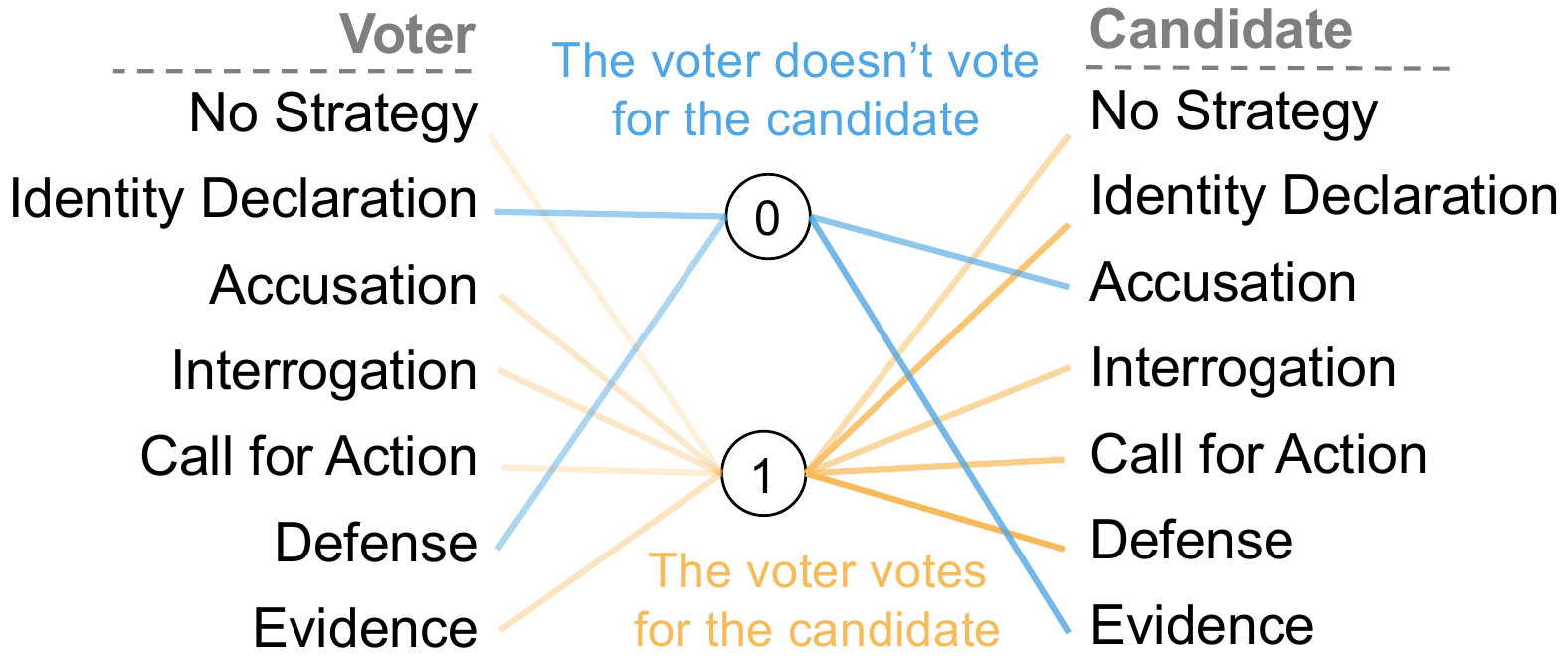}
\caption{Weights visualization of persuasion strategies in logistic regression. The connection between a strategy and 0 means this strategy contributes to the prediction of 0 (\emph{i.e.} the voter doesn't vote for the candidate). Likewise, the connection between a strategy and 1 denotes this strategy contributes to the prediction of 1 (\emph{i.e.} the voter votes for the candidate). The transparency of lines corresponds to the weights of logistic regression. A less transparent line suggests a greater weight and more impact on the output.}
\vspace{-0.5em}
\label{fig:weight_vis}
\end{figure}

We visualize the weights of logistic regression in Fig.~\ref{fig:weight_vis}. Interestingly, for positive prediction (the voter votes for the candidate), the weights of the candidate are higher than the voter. It indicates that whether the vote happens between two players depends more on the candidate's behaviors. This conforms to our intuition that players make their decisions relying on candidates' arguments.

As for the negative prediction, the evidence is the most important strategy for the candidate to clear up the suspicion. It confirms that players incline to trust those who provide more information and evidence to find the werewolf.

\vspace{-0.3em}
\section{Conclusion and Future Work}
In this work, we introduce the first persuasion modeling dataset with multiple modalities and rich utterance-level persuasion strategy annotations. We design a computational model that leverages both textual and visual representations for understanding persuasion behaviors. Our experiments suggest that visual cues can benefit the model's performance on persuasion strategy prediction. We call for future work to explore how the audio modality may contribute to persuasion modeling. We will also study the joint learning of multimodal representation in the social persuasion setting.

\section*{Limitations}
We only use the pre-trained video transformers off-the-shelf to encode the videos, while more nuanced and specific utilization of other models can be explored to further improve the performance. There are also valuable egocentric videos and demographic statistics along with the Ego4D dataset that we haven't managed to incorporate effectively. Due to the difficulty and cost of collecting videos with transcription and voting outcome annotations, the total number of games is insufficient to train a deep neural network for voting outcome deduction, though data augmentation techniques can be explored to mitigate this limitation.

\section*{Ethics Statement}
How humans use persuasion strategies in their communication has been long studied in psychology, communication, and NLP ~\citep{hovland1953communication, crano2006attitudes, petty1986elaboration,yang-etal-2019-lets, wang-etal-2019-persuasion, chen2021weakly}. We recognize that persuasion skills could be used for good and bad purposes. In this study, our goal is to study persuasive behaviors by multiple speakers through social deduction games. While we recognize that in both games of One Night Ultimate Werewolf and Avalon games players could use persuasion strategies for behaviors that perhaps are considered morally wrong, such as deception, bias, and emotional manipulation, our study  does not encourage such behaviors. Instead, we aim to understand people's behavior in a group setting when persuasion happens. Having these persuasion skills could benefit people to perform well in their workplace, such as pitching their ideas, or advocating for peace-making~\citep{simons1976persuasion}.

For our data collection and annotation process, this study has been reviewed and approved by our institution's internal review board. We obtain consent from the players who are recorded and de-identify personally identifiable information (PII), as part of the Ego4D efforts.
Moreover, to mitigate potential risks of harmful usage of this dataset in the future, we ask any users to sign an online agreement before using our resources for their research as follows: "\textit{I will not use this dataset for malicious purposes (but not limited to): deception, impersonation, mockery, discrimination, manipulation, targeted harassment, and hate speech.}"

\bibliography{anthology,custom}
\bibliographystyle{acl_natbib}

\appendix

\section{Game Rules}
\label{app:rule}
\noindent\textbf{One Night Werewolf}.
In this game, players are divided into two teams -- the team of villagers and the team of werewolves. In each game, players close their eyes in the night phase and take some actions (\emph{e.g.} swapping cards) depending on their roles. Players' roles might be changed during the night, but they don't know their new roles except for a few special roles. Then all players open their eyes. The villager team needs to find the werewolf through communication and negotiation. The werewolf team must mislead the others and try to hide their identities. At the end of the game, everyone has to point out the most suspicious player. If at least one werewolf is voted out, the villager team wins the game. Otherwise, the werewolf team wins. We refer to \url{https://en.wikipedia.org/wiki/Ultimate_Werewolf#One_Night_roles} for detailed explanations of game rules and roles.

\noindent\textbf{The Resistance: Avalon}.
In this game, players are divided into two teams -- the team of Minions and the team of Loyal Servants of Arthur.  After shuffling and distributing cards to players, they secretly check their role cards and place them face down on the table. Each player will take turns serving as the Leader. In each round, the Leader proposes a Team to do a Quest, and all players are involved in discussing if the Team assignment is passed or rejected. After the Team Building phase, the approved Team will decide if the Quest is successful or not. In the Quest phase, the Good Team can only use the Quest Success card, and the Evil Team can use either Success or Fail card. The Good Team wins when three successful Quests are made, while the Evil Team wins when three failed Quests are made or the Evil players identify Merlin in the Good Team. We refer to \url{https://en.wikipedia.org/wiki/The_Resistance_(game)#Avalon_variant} for detailed explanations of game rules and roles.

\section{Transcription Interface}
\label{app:transcription_interface}
\begin{figure*}[t]
\centering
\framebox{\includegraphics[width=\linewidth]{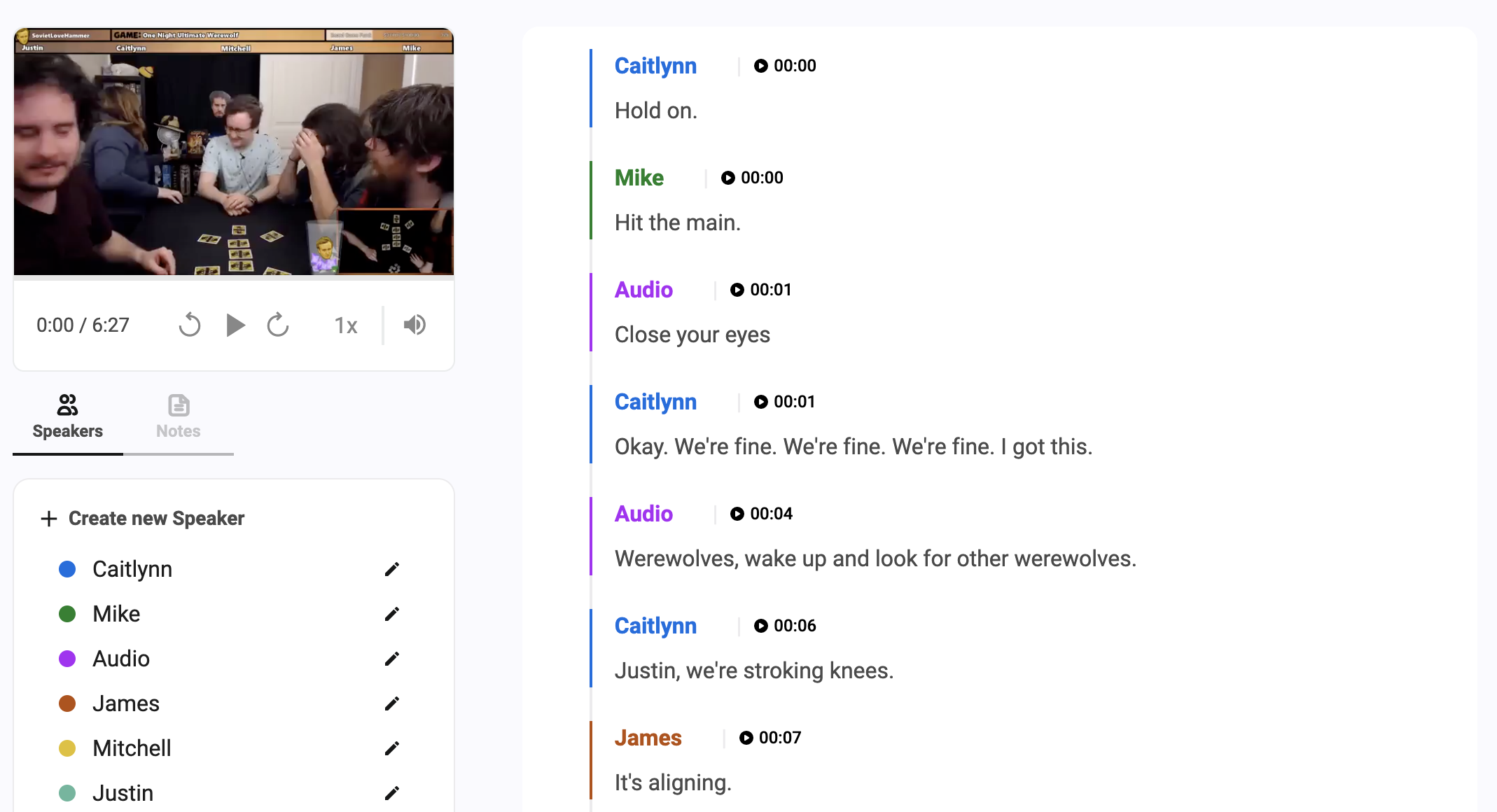}}
\caption{Screenshot of the transcription tool.}
\label{fig:transcription_tool}
\end{figure*}

We provide a screenshot of the transcription tool (\emph{rev.com}) in Fig~\ref{fig:transcription_tool}. We upload video clips to this online platform for transcription. We also provide player names and roles involved in each game to make the transcription more accurate. As illustrated in Fig~\ref{fig:transcription_tool}, they return the transcription of each utterance, the name of the speaker and the corresponding timestamp. Then we ask annotators to watch videos again and examine the alignment of videos and transcripts. Annotators also correct errors in speakers' names and texts.

\section{Annotation Interface}
\label{app:interface}
\begin{figure*}[t]
\centering
\framebox{\includegraphics[height=\dimexpr \textheight - 1\baselineskip\relax]{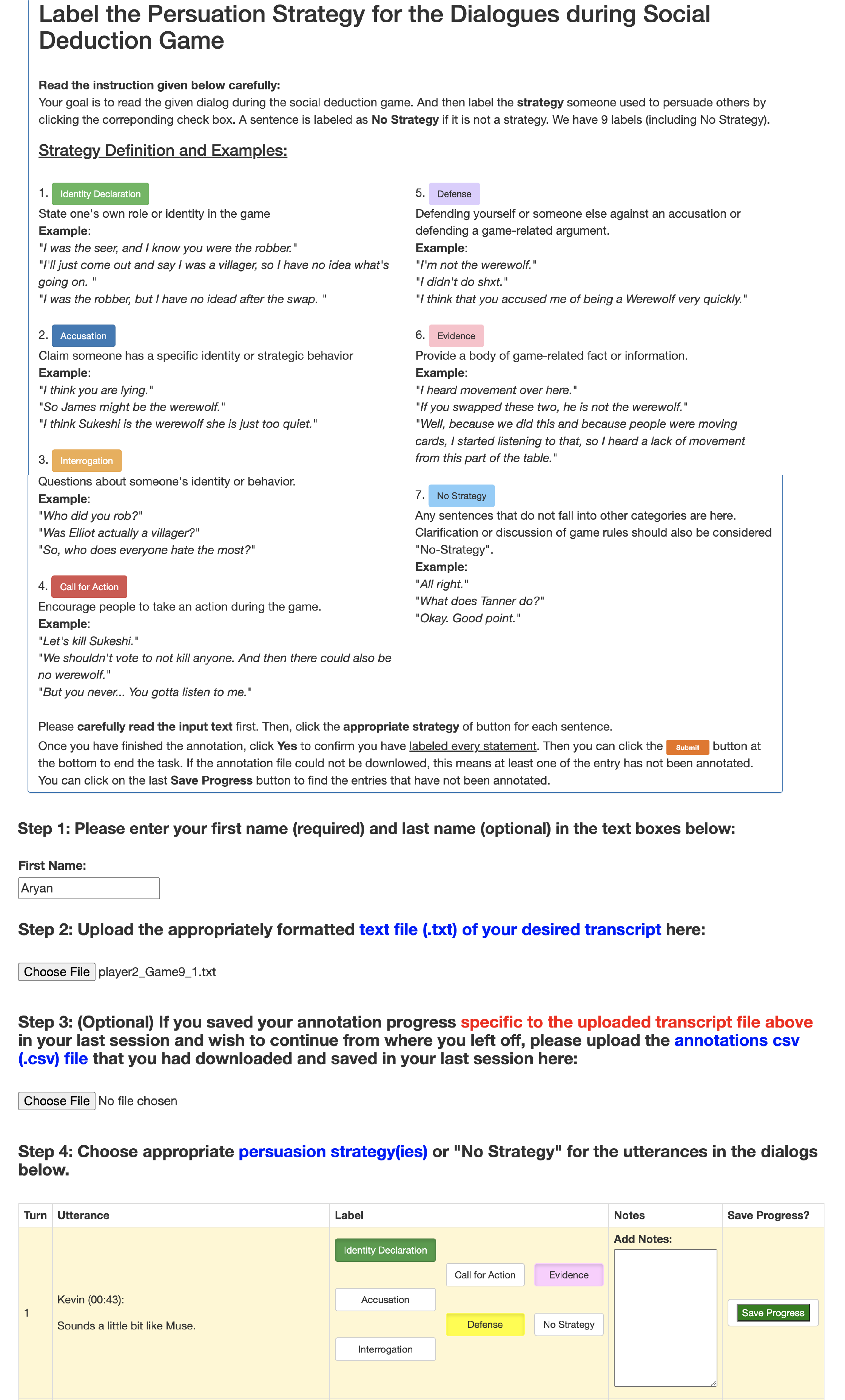}}
\caption{The screenshot of persuasion strategy annotation interface}
\label{fig:annot_interface}
\end{figure*}

We provide a screenshot in Fig.~\ref{fig:annot_interface} of our interface used by annotators to annotate utterance-level persuasion strategies.

\section{Details of Video Representation}
\label{app:video}
We now introduce the video representation extraction. Given an utterance $U_i$, We first localize the corresponding video segment using the utterance timestamp $t_i$, and then approximate the duration of the utterance by $d_i=t_{i+1}-t_i$. We have an average duration of 2 seconds on both Ego4D and Youtube datasets. To tolerate some misalignment of videos and transcripts, we set a 2-second time window for utterances shorter than 2 seconds, and hence the final duration of an utterance $U_i$ is $d_i'=\max(d_i, 2)$. Then we sample $N$ frames out of the corresponding video segment with equal spacing, \emph{i.e.}, $\mathcal{V} = \{\mathcal{V}_1, \mathcal{V}_2, \dots, \mathcal{V}_N\}$. All videos in our dataset have an aspect ratio of 16:9. Hence we make three square crops on the left, center and right of each frame to cover the entire view. Correspondingly, The visual embedding from vision encoder is composed of three parts, \emph{i.e.}, $\mathcal{V}_i = \{\mathcal{V}^{left}_i, \mathcal{V}^{center}_i, \mathcal{V}^{right}_i\}$. We input the left crops, center crops and right crops into video encoder separately and obtain the corresponding representations, $\emph{i.e.}, \psi(\mathcal{V})=\{\psi(\mathcal{V}^{left}), \psi(\mathcal{V}^{center}), \psi(\mathcal{V}^{right})\}$. The three video representations are flattened as a single vector when we concatenate them with the text representation. In our experiments, we adopt the 24-layer multiscale vision transformer (MViT)~\citep{fan2021multiscale} pretrained on Kinetics-400 as the video encoder. The number of sampled frames $N$ is set as 32 in our experiments. Note that we don't finetune the video encoder on our datasets because the performance of some models drops after finetuning due to overfitting.

In the experiments on Ego4D Werewolf data and Youtube data, the video features improve the performance prominently. However, the video feature does not necessarily help with model generalization. When we apply the model \emph{Base+V} trained on Youtube data to Ego4D data, the average F1 of BERT and RoBERTa drop by 0.8\% and 3.0\%, respectively, while the F1 of MT-BERT increases by 1.1\% after involving video features. This suggests a domain gap exists in the videos of the two datasets which is caused by the differences in the camera locations, angles of views, brightness in the room and etc. Video models are sensitive to these visual differences, resulting in limited performance in generalization on different datasets. In contrast, players communicate in a similar way in different conditions, so the pure text model generalizes better to other data.

\section{Per-class Results for Experiments with Different Context Lengths}
We showcase our experimental results including per-strategy scores on incorporating additional conversational context, in full detail in Table~\ref{table:context_results}.
\label{app:per-class}

\begin{table*}[t]
\footnotesize 
\centering
{
\setlength{\tabcolsep}{2pt}
\setlength{\extrarowheight}{2pt}

\begin{tabular}{c|c|cccccc|cc}
\toprule
&Method &Identity	&Accusation	&Interrogation	&Call for Action	&Defense	&Evidence	&Avg F1	&Joint-A       \\ 
\hline 
\multirow{12}{*}{\rotatebox[origin=c]{90}{Ego4D}}  
&\makecell{BERT}  & {\fontseries{b}\selectfont 82.6}\ci{1.1}  & $48.8$\ci{4.8}  & $82.8$\ci{0.2}  & $39.4$\ci{9.6}  & $29.3$\ci{5.5}  & $54.2$\ci{2.5}  & $56.2$\ci{2.5}  & $65.1$\ci{1.6}  \\
&\makecell{BERT + context1} & $80.4$\ci{0.8}  & $49.0$\ci{3.6}	 & $82.6$\ci{0.4}  & $46.4$\ci{8.3}	 & $29.1$\ci{2.8}  & $53.8$\ci{3.6}	 & $56.9$\ci{0.7}  & $64.0$\ci{0.9}  \\
&\makecell{BERT + context3} & $81.7$\ci{1.3}  & $51.8$\ci{4.7}	 & $81.1$\ci{2.6}  & $45.7$\ci{4.6}	 & $32.5$\ci{2.4}  & $52.2$\ci{0.9}	 & $57.5$\ci{1.6}  & $64.8$\ci{1.0}  \\
&\makecell{BERT + context5} & $79.9$\ci{1.6}  & $52.0$\ci{3.3}	 & $81.0$\ci{1.1}  & $49.5$\ci{3.2}	 & $33.8$\ci{0.5}  & $57.1$\ci{1.6}	 & $58.9$\ci{0.6}  & $65.0$\ci{0.2}  \\
&\makecell{BERT + context7} & $80.7$\ci{3.1}  & $47.4$\ci{5.4}	 & $80.7$\ci{1.7}  & $38.6$\ci{12.0}	 & $34.7$\ci{2.2}  & $55.3$\ci{0.7}	 & $56.3$\ci{2.4}  & $63.5$\ci{1.0}  \\
&\makecell{BERT + context9} & $77.9$\ci{0.1}  & $47.5$\ci{2.7}	 & $78.5$\ci{1.9}  & $43.0$\ci{5.3}	 & $31.8$\ci{0.9}  & $54.1$\ci{3.5}	 & $55.5$\ci{0.7}  & $63.7$\ci{0.9}  \\
\cdashline{2-10}
&\makecell{RoBERTa}   & $81.7$\ci{2.6}  & $51.7$\ci{0.9}  & $83.4$\ci{0.9}  & $43.3$\ci{8.7}  & $33.1$\ci{2.2}  & $51.7$\ci{2.1}  & $57.5$\ci{1.4}  & $63.4$\ci{0.5}  \\
&\makecell{RoBERTa + context1} & $79.9$\ci{2.8}  & $53.1$\ci{0.6}	 & $82.1$\ci{0.9}  & $41.8$\ci{7.9}	 & $34.1$\ci{1.4}  & $55.2$\ci{2.9}	 & $57.7$\ci{1.4}  & $64.1$\ci{0.7}  \\
&\makecell{RoBERTa + context3} & $81.7$\ci{1.0}  & $53.9$\ci{3.9}	 & $82.3$\ci{1.6}  & $39.6$\ci{9.0}	 & $35.4$\ci{2.9}  & $54.0$\ci{3.8}	 & $57.8$\ci{2.1}  & $64.1$\ci{1.5}  \\
&\makecell{RoBERTa + context5} & $81.5$\ci{0.7}  & {\fontseries{b}\selectfont 59.4}\ci{2.4}  & $83.5$\ci{1.1}  & $43.7$\ci{3.7}  & $33.0$\ci{3.1}  & $52.4$\ci{2.9}  & $58.9$\ci{1.2}  & $64.6$\ci{0.7}   \\
&\makecell{RoBERTa + context7} & $78.6$\ci{1.7}  & $55.5$\ci{0.5}	 & $80.6$\ci{0.4}  & $38.2$\ci{4.0}	 & $30.1$\ci{4.9}  & $51.9$\ci{3.2}	 & $55.8$\ci{1.2}  & $62.4$\ci{2.3}  \\
&\makecell{RoBERTa + context9} & $80.2$\ci{1.8}  & $56.0$\ci{2.2}	 & $83.0$\ci{1.4}  & $42.5$\ci{10.4}	 & $32.0$\ci{2.4}  & $53.5$\ci{1.7}	 & $57.9$\ci{1.8}  & $63.0$\ci{0.6}  \\\cdashline{2-10}
&\makecell{MT-BERT} & $80.9$\ci{1.3}  & $51.5$\ci{3.3}  & $83.0$\ci{1.3}  & {\fontseries{b}\selectfont 56.6}\ci{2.3}  & $25.9$\ci{2.0}  & $53.6$\ci{1.3}  & $58.6$\ci{0.3}  & $65.5$\ci{0.8} \\
&\makecell{MT-BERT + context1} & $79.2$\ci{2.2}  & $53.3$\ci{2.3}	 & {\fontseries{b}\selectfont 84.3}\ci{0.6}  & $52.9$\ci{2.9}	 & $31.1$\ci{5.8}  & $55.0$\ci{3.2}	 & $59.3$\ci{0.4}  & $66.0$\ci{2.0}  \\
&\makecell{MT-BERT + context3} & $77.4$\ci{2.2}  & $52.6$\ci{3.8}	 & $83.2$\ci{2.1}  & $46.2$\ci{2.4}	 & $35.1$\ci{2.3}  & $56.1$\ci{2.7}	 & $58.4$\ci{0.1}  & $65.1$\ci{0.7}  \\
&\makecell{MT-BERT + context5}    & $79.8$\ci{2.2}  & $54.4$\ci{0.8}  & $83.2$\ci{0.7}  & $50.8$\ci{7.2}	 &{\fontseries{b}\selectfont 36.5}\ci{2.8}  & {\fontseries{b}\selectfont 61.5}\ci{2.2}	 & {\fontseries{b}\selectfont 61.0}\ci{1.1}  & {\fontseries{b}\selectfont 66.3}\ci{1.4}  \\
&\makecell{MT-BERT + context7} & $78.5$\ci{2.5}  & $54.7$\ci{3.3}	 & $82.6$\ci{1.2}  & $47.9$\ci{2.5}	 & $33.5$\ci{2.2}  & $53.4$\ci{1.4}	 & $58.4$\ci{1.1}  & $65.0$\ci{0.9}  \\
&\makecell{MT-BERT + context9} & $78.2$\ci{2.2}  & $54.7$\ci{1.6}	 & $82.1$\ci{0.4}  & $47.8$\ci{3.8}	 & $30.5$\ci{5.8}  & $56.3$\ci{1.0}	 & $58.3$\ci{0.5}  & $64.8$\ci{0.8}  \\
\hline
\multirow{12}{*}{\rotatebox[origin=c]{90}{YouTube}}  
&\makecell{BERT}   & $80.2$\ci{1.6}  & $64.7$\ci{1.1}  & $89.6$\ci{0.4}  & $77.2$\ci{2.5}  & $43.5$\ci{1.0}  & $58.3$\ci{0.7}  & $68.9$\ci{0.1}  & $64.6$\ci{0.8}  \\
&\makecell{BERT + context1} & $81.2$\ci{1.1}  &  $66.5$\ci{0.5}	 & $90.2$\ci{0.3}  & $77.7$\ci{0.3}	 & $43.6$\ci{2.7}  & $59.5$\ci{0.6}	 & $69.8$\ci{0.3}  & $65.1$\ci{0.8}  \\
&\makecell{BERT + context3} & $82.6$\ci{0.7}  & $65.9$\ci{0.5}	 & $90.1$\ci{0.7}  & $77.4$\ci{1.2}	 & $43.0$\ci{1.5}  & $60.4$\ci{0.9}	 & $69.9$\ci{0.4}  & $64.4$\ci{0.7}  \\
&\makecell{BERT + context5} & $82.6$\ci{0.7}  & $66.7$\ci{1.0}  & $89.6$\ci{1.5}  & $78.1$\ci{2.4}  & $45.7$\ci{1.1}  & $59.7$\ci{1.1}  & $70.4$\ci{0.3}  & $64.4$\ci{1.0}  \\
&\makecell{BERT + context7} & $81.8$\ci{0.6}  & $67.2$\ci{1.2}	 & $90.5$\ci{0.2}  & $77.7$\ci{0.5}	 & $45.0$\ci{0.7}  & $60.2$\ci{1.2}	 & $70.4$\ci{0.5}  & $64.8$\ci{1.0}  \\
&\makecell{BERT + context9} & $80.6$\ci{1.1}  & $66.7$\ci{0.4}	 & $90.3$\ci{0.2}  & $77.0$\ci{1.2}	 & $42.2$\ci{2.0}  & $59.6$\ci{0.2}	 & $69.4$\ci{0.4}  & $64.0$\ci{1.3}  \\
\cdashline{2-10}
&\makecell{RoBERTa}   &  {\fontseries{b}\selectfont 84.3}\ci{0.1}  & $67.2$\ci{0.6}  & $89.4$\ci{0.1}  & $78.2$\ci{0.8}  & $44.3$\ci{0.4}  & $59.0$\ci{1.7}  & $70.4$\ci{0.2}  & $64.8$\ci{0.7} \\
&\makecell{RoBERTa + context1} & $83.3$\ci{0.2}  & $67.0$\ci{0.3}	 & $89.9$\ci{0.2}  &  {\fontseries{b}\selectfont 78.4}\ci{0.9}	 & $43.4$\ci{2.7}  & $59.7$\ci{0.5}	 & $70.3$\ci{0.5}  & $65.7$\ci{1.0}  \\
&\makecell{RoBERTa + context3} & $82.7$\ci{1.5}  &  {\fontseries{b}\selectfont 67.8}\ci{0.1}	 & $90.3$\ci{0.4}  & $77.4$\ci{0.4}	 & $43.1$\ci{1.5}  &  {\fontseries{b}\selectfont 61.0}\ci{1.9}	 & $70.4$\ci{0.2}  & $65.5$\ci{0.4}  \\
&\makecell{RoBERTa + context5} & $82.4$\ci{0.3}  & $67.0$\ci{1.1}  & $90.2$\ci{0.0}  & $77.1$\ci{1.0}  & $46.1$\ci{0.7}  & $59.9$\ci{0.7}  &  {\fontseries{b}\selectfont 70.5}\ci{0.3}  & $64.7$\ci{0.6}  \\
&\makecell{RoBERTa + context7} & $83.5$\ci{0.8}  & $66.0$\ci{0.6}	 & $90.2$\ci{0.4}  & $77.8$\ci{0.3}	 & $46.6$\ci{1.5}  & $58.4$\ci{1.1}	 & $70.4$\ci{0.3}  & $65.1$\ci{1.3}  \\
&\makecell{RoBERTa + context9} & $82.9$\ci{2.0}  & $66.6$\ci{0.7}	 &  {\fontseries{b}\selectfont 90.6}\ci{0.2}  & $75.5$\ci{0.5}	 &  {\fontseries{b}\selectfont 46.8}\ci{0.9}  & $58.9$\ci{1.1}	 & $70.2$\ci{0.3}  & $64.9$\ci{0.9}  \\
\cdashline{2-10}
&\makecell{MT-BERT} & $80.7$\ci{0.4}  & $65.1$\ci{1.5}  & $88.5$\ci{0.8}  & $76.2$\ci{2.2}  & $42.3$\ci{1.5}  & $57.4$\ci{1.3}  & $68.4$\ci{0.3}  & $65.6$\ci{1.1} \\
&\makecell{MT-BERT + context1} & $82.9$\ci{0.9}  & $67.2$\ci{1.2}	 & $88.7$\ci{1.5}  & $77.8$\ci{1.6}	 & $43.4$\ci{0.6}  & $59.0$\ci{0.8}	 & $69.8$\ci{0.3}  & $67.3$\ci{0.9}  \\
&\makecell{MT-BERT + context3} & $80.5$\ci{2.5}  & $65.9$\ci{1.5}	 & $89.9$\ci{0.2}  & $75.2$\ci{1.4}	 & $44.9$\ci{1.9}  & $58.3$\ci{0.4}	 & $69.1$\ci{0.6}  & $65.8$\ci{0.9}  \\
&\makecell{MT-BERT + context5} & $83.1$\ci{1.1}  & $65.0$\ci{1.5}  & $90.1$\ci{0.3}  & $74.6$\ci{2.4} & $46.5$\ci{0.8}  & $59.2$\ci{0.3}  & $69.7$\ci{0.6}  & $66.7$\ci{0.5} \\
&\makecell{MT-BERT + context7} & $82.1$\ci{0.8}  & $67.3$\ci{0.1}	 & $89.4$\ci{1.1}  & $76.0$\ci{0.9}	 & $43.2$\ci{1.3}  & $57.8$\ci{0.5}	 & $69.3$\ci{0.3}  &  {\fontseries{b}\selectfont 67.6}\ci{0.7}  \\
&\makecell{MT-BERT + context9} & $81.0$\ci{2.0}  & $67.8$\ci{0.2}	 & $89.6$\ci{0.6}  & $72.8$\ci{3.8}	 & $44.3$\ci{2.1}  & $58.8$\ci{1.0}	 & $69.1$\ci{0.8}  & $66.5$\ci{0.5}  \\
\bottomrule
\end{tabular}}
\caption{Experimental Results on incorporating the conversational context of different lengths for persuasion strategy prediction}
\label{table:context_results}
\end{table*}

\section{Experiments of Domain Generalization}
\label{app:domain_generalization}
We demonstrate the detailed experiment results of data domain generalization (training models on Youtube data and testing on Ego4D Werewolf test set), as well as game domain generalization (training models on Ego4D Werewolf data and testing on Avalon data). Results are reported in Table~\ref{table:data_generalization} and Table~\ref{table:game_generalization}, respectively.

\begin{table*}[t]
\footnotesize 
\centering
{
\setlength{\tabcolsep}{3.2pt}
\setlength{\extrarowheight}{2pt}

\begin{tabular}{c|c|cccccc|cc}
\toprule
&Method &Identity	&Accusation	&Interrogation	&Call for Action	&Defense	&Evidence	&Avg F1	&Joint-A       \\ 
\hline 

\multirow{6}{*}{\rotatebox[origin=c]{90}{w.o. Fine Tuning}}  
&\makecell{BERT}   & $82.0$\ci{1.2}  & $53.9$\ci{1.6}  & $84.1$\ci{0.9}  & $53.0$\ci{4.1}  & $33.9$\ci{0.5}  & $53.5$\ci{3.9}  & $60.1$\ci{0.7}  & $65.6$\ci{1.0}   \\
&\makecell{BERT + C}   & $83.6$\ci{0.8}  & $55.7$\ci{1.1}  & $85.6$\ci{1.0}  & $46.5$\ci{2.7}  & $34.5$\ci{2.5}  & $60.9$\ci{3.2}  & $61.1$\ci{0.9}  & $65.6$\ci{1.3}  \\
\cdashline{2-10}
&\makecell{RoBERTa}  & $86.9$\ci{0.9}  & $57.0$\ci{1.4}  & $85.0$\ci{2.0}  & $53.5$\ci{3.6}  & $31.5$\ci{1.3}  & $55.0$\ci{1.3}  & $61.5$\ci{0.8}  & $66.0$\ci{0.7} \\
&\makecell{RoBERTa + C}  & $82.5$\ci{2.4}  & $56.3$\ci{2.7}  & $86.2$\ci{0.6}  & $50.7$\ci{4.1}  & $37.6$\ci{1.8} & $59.6$\ci{1.8}  & $62.2$\ci{1.3}  & $67.6$\ci{0.6} \\
\cdashline{2-10}
&\makecell{MT-BERT} & $80.6$\ci{2.9}  & $50.4$\ci{3.6}  & $83.5$\ci{1.7}  & $45.3$\ci{10.6}  & $34.7$\ci{0.7}  & $55.2$\ci{3.2}  & $58.3$\ci{2.7}  & $64.8$\ci{2.1}  \\
&\makecell{MT-BERT + C}  & $81.8$\ci{2.1}  & $53.8$\ci{2.7}  & $83.4$\ci{1.9}  & $44.1$\ci{8.2}  & $35.9$\ci{1.7}  & $53.5$\ci{2.1}  & $58.7$\ci{2.3}  & $66.3$\ci{2.0} \\
\hline
\multirow{6}{*}{\rotatebox[origin=c]{90}{w. Fine Tuning}}  
&\makecell{BERT}  & $82.0$\ci{1.4}  & $54.9$\ci{0.9}  & $82.8$\ci{0.6}  & $53.0$\ci{1.9}  & $29.9$\ci{1.0}  & $61.7$\ci{0.7}  & 60.7\ci{0.2}  & $68.1$\ci{0.2} \\
&\makecell{BERT + C}  & $84.1$\ci{0.1}  & $55.6$\ci{3.5}  & $86.1$\ci{0.4}  & $49.9$\ci{2.2}  & $32.6$\ci{1.5}  & $61.5$\ci{2.4}  & $61.6$\ci{1.1}  & $69.0$\ci{0.8} \\
\cdashline{2-10}
&\makecell{RoBERTa}  & $86.7$\ci{1.3}  & $56.6$\ci{1.4}  & $85.3$\ci{1.6}  & $54.8$\ci{3.9}  & $29.4$\ci{2.5}  & $57.3$\ci{2.0}  & $61.7$\ci{1.4}  & $67.4$\ci{0.9} \\
&\makecell{RoBERTa + C}   & $84.0$\ci{1.5}  & $58.9$\ci{1.6}  & $84.9$\ci{0.2}  & $52.4$\ci{4.4}  & $38.0$\ci{2.3}  & $62.5$\ci{2.4}  & $63.4$\ci{1.7}  & $69.1$\ci{0.6} \\
\cdashline{2-10}
&\makecell{MT-BERT}  & $81.9$\ci{1.4}  & $54.7$\ci{1.6}  & $83.0$\ci{0.6}  & $60.2$\ci{5.3}  & $25.6$\ci{1.6}  & $59.2$\ci{2.3}  & $60.8$\ci{1.0}  & $68.5$\ci{0.6} \\
&\makecell{MT-BERT + C}  & $83.7$\ci{0.9}  & $54.3$\ci{2.4}  & $84.5$\ci{1.1}  & $53.8$\ci{3.2}  & $33.8$\ci{3.4}  & $58.1$\ci{2.8}  & $61.4$\ci{1.0}  & $70.0$\ci{1.5} \\
\bottomrule
\end{tabular}}
\caption{Data domain generalization experiments. We train models on Youtube data and test on Ego4D testing set. Then we fine-tune the models on Ego4D training set and test again.}
\label{table:data_generalization}
\end{table*}

\begin{table*}[t]
\footnotesize 
\centering
{
\setlength{\tabcolsep}{3.2pt}
\setlength{\extrarowheight}{2pt}

\begin{tabular}{c|cccccc|cc}
\toprule
Method &Identity	&Accusation	&Interrogation	&Call for Action	&Defense	&Evidence	&Avg F1	&Joint-A       \\ 
\hline 
\makecell{BERT}   & $66.1$\ci{3.7}  & $37.2$\ci{4.9}  & $73.3$\ci{3.7}  & $20.9$\ci{7.3}  & $23.5$\ci{6.3}  & $26.6$\ci{5.3}  & $41.3$\ci{3.3}  & $62.9$\ci{0.9}   \\
\makecell{BERT + C}   & $60.4$\ci{4.5}  & $41.7$\ci{2.2}  & $73.2$\ci{0.3}  & $34.4$\ci{4.8}  & $25.0$\ci{4.3}  & $18.2$\ci{2.9}  & $42.1$\ci{1.4}  & $63.9$\ci{0.6}  \\
\cdashline{1-9}
\makecell{RoBERTa}  & $63.0$\ci{3.2}  & $45.9$\ci{3.0}  & $73.1$\ci{2.8}  & $33.0$\ci{12.1}  & $33.6$\ci{5.7}  & $27.5$\ci{1.2}  & $46.0$\ci{1.8}  & $64.3$\ci{0.5} \\
\makecell{RoBERTa + C}  & $46.1$\ci{9.3}  & $40.3$\ci{2.3}  & $71.8$\ci{2.9}  & $35.8$\ci{2.8}  & $28.1$\ci{3.0}  & $22.5$\ci{8.5}  & $40.8$\ci{1.7}  & $63.5$\ci{0.6} \\
\cdashline{1-9}
\makecell{MT-BERT} & $56.3$\ci{3.6}  & $38.2$\ci{4.6}  & $73.1$\ci{1.9}  & $39.3$\ci{5.7}  & $23.5$\ci{8.0}  & $25.3$\ci{1.0}  & $42.6$\ci{1.7}  & $63.4$\ci{0.2}  \\
\makecell{MT-BERT + C}  & $61.9$\ci{2.9}  & $42.2$\ci{2.5}  & $71.2$\ci{2.7}  & $34.2$\ci{1.0} & $29.3$\ci{5.2}  & $23.2$\ci{3.7}  & $43.7$\ci{1.2}  & $64.2$\ci{0.3} \\
\bottomrule
\end{tabular}}
\caption{Game domain generalization experiments. We train models on Ego4D Werewolf training set and test them on Avalon data.}
\label{table:game_generalization}
\end{table*}

\section{Detailed prompt templates used for GPT-3}
\label{app:prompts}
For the prompt templates, we use the guideline and the persuasion strategy definitions provided to the annotators under the zero-shot setting, and append one/five more examples under one/five-shot setting. The detailed prompt template we used for GPT-3 inference is shown in Table~\ref{tab:prompt}.

\begin{table*}[]
    \centering
    \scalebox{0.93}{
    \begin{tabular}{lp{0.93\textwidth}}
    \toprule
zero-shot &
Label the Persuasion Strategy for the Utterances in Dialogues during Social Deduction Game. Do not hesitate to select multiple strategies if one category can not summarize the given utterance.
         
\bigskip

Strategy Definition:

\bigskip

1. Identity Declaration: State one's own role or identity in the game

2. Accusation: Claim someone has a specific identity or strategic behavior

3. Interrogation: Questions about someone's identity or behavior

4. Call for Action: Encourage people to take an action during the game

5. Defense: Defending yourself or someone else against an accusation or defending a game-related argument

6. Evidence: Provide a body of game-related facts or information

7. No Strategy: Any sentences that do not fall into other categories are here. Clarification or discussion of game rules should also be considered "No-Strategy"

\bigskip

Utterance: "\$utterance\$"

Strategy:
         \\
    \midrule
one-shot &
Label the Persuasion Strategy for the Utterances in Dialogues during Social Deduction Game. Do not hesitate to select multiple strategies if one category can not summarize the given utterance.
         
\bigskip

Strategy Definition:

\emph{[same as above]}

\bigskip

Utterance: "No, but in order to find it, I had to really tap around to find it."

Strategy: Defense, Evidence

\bigskip

Utterance: "\$utterance\$"

Strategy:
         \\
    \midrule
five-shot &
Label the Persuasion Strategy for the Utterances in Dialogues during Social Deduction Game. Do not hesitate to select multiple strategies if one category can not summarize the given utterance.
         
\bigskip

Strategy Definition:

\emph{[same as above]}

\bigskip

Utterance: "I'll just come out and say I was a villager, so I have no idea what's going on."

Strategy: Identity Declaration

\bigskip

Utterance: "So James might be the werewolf."

Strategy: Accusation

\bigskip

Utterance: "Did anybody do any swapping? Anybody willing to fess up to anything about swapping?"

Strategy: Interrogation, Call for Action

\bigskip

Utterance: "No, but in order to find it, I had to really tap around to find it."

Strategy: Defense, Evidence

\bigskip

Utterance: "Okay. Good point."

Strategy: No Strategy

\bigskip

Utterance: "\$utterance\$"

Strategy:
         \\
    \bottomrule
    \end{tabular}
    }
    
    \caption{Prompt templates used for GPT-3, the variable within dollars is to be replaced with the corresponding value.}
    \label{tab:prompt}
\end{table*}

\section{Persuasion Strategy Annotation on Avalon Games}
\begin{figure*}[t]
\centering
\includegraphics[width=0.9\linewidth]{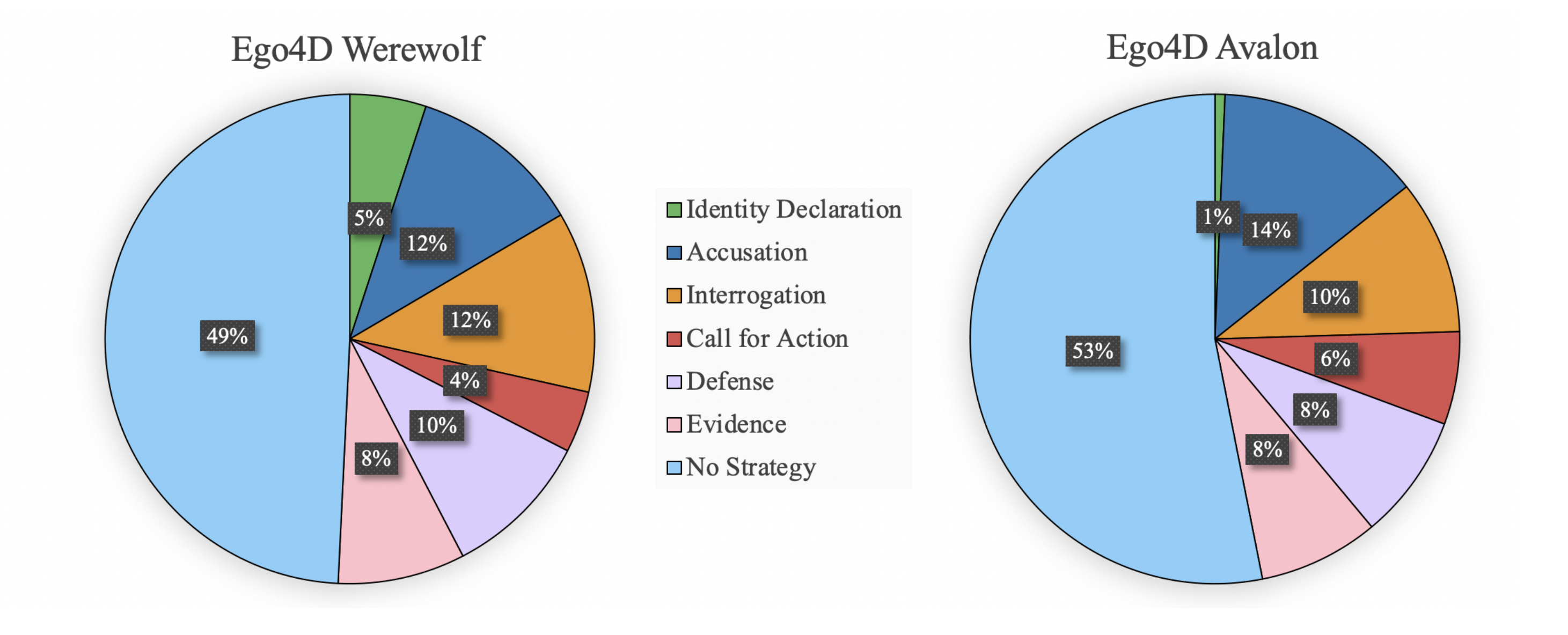}
\caption{Persuasion strategy distributions for Ego4D Werewolf games and Ego4D Avalon games: Pie charts representing the percentage of annotated utterances for every persuasion strategy and "No Strategy" in One Night Ultimate Werewolf games and The Resistance: Avalon games in Ego4D.}
\label{fig:avalon_v_werewolf}
\end{figure*}

Adjacent pie-charts comparing the distributions of annotated utterance-level persuasion strategies for One Night Ultimate Werewolf games and Avalon games in Ego4D are shown in Fig.~\ref{fig:avalon_v_werewolf}. We can observe a different distribution of adopted persuasion strategies in the two games, suggesting a large game domain gap.

\label{app:avalon}

\end{document}